\documentclass{article}

\usepackage{microtype}
\usepackage{graphicx}
\usepackage{subfigure}
\usepackage{booktabs} % for professional tables

\usepackage{hyperref}

\usepackage[accepted]{icml2024}

\usepackage{amssymb}
\usepackage{mathtools}
\usepackage{amsthm}

\usepackage[capitalize,noabbrev]{cleveref}

\theoremstyle{plain}
\newtheorem{theorem}{Theorem}[section]

\newtheorem{lemma}[theorem]{Lemma}
\newtheorem{corollary}[theorem]{Corollary}
\theoremstyle{definition}
\newtheorem{definition}[theorem]{Definition}
\newtheorem{assumption}[theorem]{Assumption}
\theoremstyle{remark}
\newtheorem{remark}[theorem]{Remark}

\usepackage[textsize=tiny]{todonotes}

\icmltitlerunning{Can Machines Learn True Probabilities?}

\begin{document}

\twocolumn[
\icmltitle{Can Machines Learn the True Probabilities?}

% It is OKAY to include author information, even for blind
% submissions: the style file will automatically remove it for you
% unless you've provided the [accepted] option to the icml2024
% package.

% List of affiliations: The first argument should be a (short)
% identifier you will use later to specify author affiliations
% Academic affiliations should list Department, University, City, Region, Country
% Industry affiliations should list Company, City, Region, Country

% You can specify symbols, otherwise they are numbered in order.
% Ideally, you should not use this facility. Affiliations will be numbered
% in order of appearance and this is the preferred way.
%\icmlsetsymbol{}{*}

\begin{icmlauthorlist}
\icmlauthor{Jinsook Kim}{Yonsei}
%\icmlauthor{Firstname2 Lastname2}{equal,yyy,comp}
%\icmlauthor{Firstname3 Lastname3}{comp}
%\icmlauthor{Firstname4 Lastname4}{sch}
%\icmlauthor{Firstname5 Lastname5}{yyy}
%\icmlauthor{Firstname6 Lastname6}{sch,yyy,comp}
%\icmlauthor{Firstname7 Lastname7}{comp}
%\icmlauthor{}{sch}
%\icmlauthor{Firstname8 Lastname8}{sch}
%\icmlauthor{Firstname8 Lastname8}{yyy,comp}
%\icmlauthor{}{sch}
%\icmlauthor{}{sch}
\end{icmlauthorlist}

\icmlaffiliation{Yonsei}{Underwood International College, Yonsei University, Seoul, Korea}
%\icmlaffiliation{comp}{Company Name, Location, Country}
%\icmlaffiliation{sch}{School of ZZZ, Institute of WWW, Location, Country}

\icmlcorrespondingauthor{Jinsook Kim}{jki76364@gmail.com}
%\icmlcorrespondingauthor{Firstname2 Lastname2}{first2.last2@www.uk}

% You may provide any keywords that you
% find helpful for describing your paper; these are used to populate
% the "keywords" metadata in the PDF but will not be shown in the document
\icmlkeywords{Foundations of Artificial Intelligence, Probabilistic Machine Learning, Non-parametric Estimation, Effective Calculation by Turing Machine, True Guarantee of Well-Calibration}

\vskip 0.3in
]

% this must go after the closing bracket ] following \twocolumn[ ...

% This command actually creates the footnote in the first column
% listing the affiliations and the copyright notice.
% The command takes one argument, which is text to display at the start of the footnote.
% The \icmlEqualContribution command is standard text for equal contribution.
% Remove it (just {}) if you do not need this facility.

\printAffiliationsAndNotice{}  % leave blank if no need to mention equal contribution
%\printAffiliationsAndNotice{\icmlEqualContribution} % otherwise use the standard text.

\begin{abstract}
When there exists uncertainty, AI machines are designed to make decisions so as to reach the best expected outcomes. Expectations are based on true facts about the objective environment the machines interact with, and those facts can be encoded into AI models in the form of true objective probability functions. Accordingly, AI models involve probabilistic machine learning in which the probabilities should be objectively interpreted. We prove under some basic assumptions when machines can learn the true objective probabilities, if any, and when machines cannot learn them. 
\end{abstract}

\section{Introduction}

In the standard AI model under uncertainty, how to measure the degree of uncertainty matters. This paper is about treating such measures in the form of probabilities. In particular, we focus on the true objective probabilities, if any. There are various probabilistic contexts in which the true objective probabilities matter. For example, causal relations of physical events are widely regarded as objective features of the world. Therefore, when causal relations are to be understood in terms of probabilities mainly due to various regularity issues, a probabilistic causal model should include an objective probability function that measures the true objective values about our world.

This paper addresses the question of whether machines can \textit{learn} the true objective probabilities from the data to perform such probabilistic reasoning. Under some basic assumptions, we prove that machines can learn the true objective probabilities if and only if the probabilities are directly observable by them. Roughly speaking, a true probability is directly observable by a machine when it can calculate the probability by the empirical frequency of a true population given to it. 

The outline of the proof is as follows. After defining some main concepts, we identify the Success Criterion and the necessary condition for any machine to learn the true objective probabilities. From these conditions, we derive the theorem that learning implies the true guarantee of well-calibration. Roughly speaking, ``truly guaranteed well-calibration'' means the following: when a machine collects data according to its subjective forecast along a stochastic path in which the associated events occur, the empirical frequency of the collected data matches the very probabilistic forecast of the machine with the true probability $P$- one. Now that the machine forecasts must indeed be true when the machine learns the true probabilities, this calibration property can then be understood as a calibration version of the strong law of large numbers without the independence assumption. 

Note that there exist connections here among machine forecasting, well-calibration, and machine learning. While proving our theorems, therefore, we establish connections between the true guarantee of well-calibration and various settings of the real forecasting games between Nature and a machine. In this game, what Nature forecasts are the true objective probabilities, while what the machine forecasts are its own subjective probabilities. The machine loses when Nature deviates from the probabilistic forecasts of the machine. Bridged by the property of truly guaranteed well-calibration, we then prove whether the machine learns the true probabilities or not under various settings of forecasting games.  

With this proof, we provide the fundamental scope and limit of learning the true probabilities by AI machines. One important implication is that machines can relax the independent assumption among data to learn the true probabilities but cannot relax the assumption of identical distribution such as stationarity or ergodicity along a stochastic path where any associated events occur. Another implication is to show that the problem of computability is directly connected to the problem of complexity in the case of learning the true probabilities. 

\section{Notations and Definitions}
In this section, we define some main concepts, including ``machine learning'' and ``true objective probability''. Adopting terminologies from \cite{Nilsson:11} and \cite{Boolos:02}, let us first define a \textit{machine} as an artifact or device that can \textit{effectively calculate} or \textit{compute} any target function if there exist definite and explicit instructions to do so in principle. Since we focus on probability functions in this paper, we particularly mean by ``an effectively calculating or computing device'' a machine that can in principle assign a probability measure (a value of a probability function) to each state (an argument of the probability function) in a given domain, an event space of a sigma-field.

\begin{definition}
\label{def:effeccal}
 A function is \emph{effectively calculable} or \emph{computable} when there are definite and explicit instructions, following which its functional value can be calculated \emph{in principle} for any given argument. (Boolos et al. (2002))   
\end{definition}

Two things merit to be taken into account with \cref{def:effeccal}. First, this notion of effective calculation or computation is an \textit{ideal} one with no practical limits on time, expense, etc., necessary to calculate. Therefore, a proof of the limitation on effective calculation or computation of any function will imply a fundamental limit on computability that cannot be overcome by any practical real machine. Second, as \cite{Kozen:97} points out, this notion is an \textit{informal} one, something that is supposed to be captured \textit{in common} by all formalisms such as computation by Turing machines, by the $\lambda$ -calculus and by the $\mu$ -recursive method, etc. Accordingly, once we adopt this notion of effective calculation or computation to define ``learning'', we can be flexible about which formalism would be encoded as instructions to complete a given learning task. 

Now, whatever such formalism is, machines can learn only if there exist some \textit{instructions} followed by them to complete their tasks. So we can prove that it is impossible for machines to learn any target function under certain conditions in the following way: we first suppose that there exist some successful instructions to be encoded into machine programming to learn any given function under the conditions. We then show that this supposition leads to a conclusion that is impossible to satisfy. We thereby conclude that there cannot exist such instructions for the given function and, accordingly, that machines cannot learn it. This is a simple but clear way of proving the impossibility of learning without being committed to any complex procedure of constructing any formalism such as a Turing machine or $\lambda$-calculus, etc. 

\begin{definition}
\label{def:learn}
A machine \emph{learns} when it \textit{succeeds} in effectively calculating or computing a target function, if any, after processing possibly infinite amounts of data.
\end{definition}

The phenomenon of learning must be at least computational in its essence when acquired by a machine. We thus adopt the notion of computation to define what learning is in \cref{def:learn}. Inspired by the ideas of \cite{Turing:36} and \cite{Church:36}, we require that a machine be able to effectively calculate or compute a target function when the machine can learn the function. 

In addition, we add the notion of success to \cref{def:learn}, which aims to capture the role of ``learning'' as an \emph{epistemic} notion, not just a \emph{computational} one. The epistemic notion of machine learning requires two components: if a machine learns, then (\textit{i}) it must be $\textit{indeed correct}$ most of the time and (\textit{ii}) it must be $\textit{self-assured to be correct}$ most of the time.

Learning is the phenomenon of \emph{knowledge} acquisition. Once something is learned, knowledge about it is acquired. Now, knowledge must be a true representation, and it must be so not just by luck. We thus require that (\textit{i}) what is effectively calculated or computed by a machine be \textit{true} and further that it be \textit{true most of the time} out of infinite opportunities to learn. In addition, if the machine admits errors too many times, say infinitely often, it cannot be said to learn. We thus require also that (\textit{ii}) the machine be \textit{self-assured} that what it calculates is correct most of the time. In sum, we provide the following \textbf{Success Criterion}:

 (1) If a machine achieves computational success by learning, what it acquires in the end must be true to our world most of the time, which must be assured to the machine itself. 
 
 If what the machine computes turns out to be wrong or it admits errors repeatedly too often out of infinite opportunities to learn, then its computation cannot be considered successful. Later, we prove that the Success Criterion (1) is sufficient for learning in the case of computing \emph{true probabilities} by \cref{cor:successcriterion}. We also clarify there what we mean by ``most of the time.''

\begin{definition}
\label{def:truepro}
A \emph{true probability} is what collectively constitutes a probability space, a triple $(\Omega, \mathcal{F},p)$ of random variables $S_{t}$'s in a joint true probability $p$ of the stochastic process according to which Nature generates a sequence of actual data $s_{t}$'s and each of these data is realized as such with the very true probability $P$.    
\end{definition}

Consider an enumerable set $\Omega_{t}$ of $\omega_{i}$'s called
\textit{states} at time $t$ with $t\in
\mathbb{N}$. For example, $\Omega_{t}$ may be the set
$\{\omega_{s},\omega_{c},\omega_{r}\}$ where $\omega_{s}$ denotes the state of
sunny day, $\omega_{c}$ the state of cloudy day and $\omega_{r}$ the state of
rainy day at date $t$. Also, consider the set $\Omega$ that consists of all
the infinite sequences with a representative sequence $\omega=(S_{0}%
^{-1}(s_{0}),S_{1}^{-1}(s_{1}),S_{2}^{-1}(s_{2}),\ldots).$ Here, $S_{t}%
(\omega_{i})$ is a random variable which has some numerical value $s_{t}\in
\Re$ according as which $\omega_{i}$'s are \textit{realized} at time $t$ in \textit{our world}. Now, $S_{t}$ comes before $S_{t+1}$ in time, and thus the sequence of $S_{t}$'s represents a discrete-time stochastic process. Then Nature generates the \textit{actual} data set
$\{s_{0},s_{1},s_{2},\ldots\}$ with true probability $P$'s. So the probability function $P$, if any, becomes \textit{true} to \textit{our world} when it corresponds to whatever amounts to the rules according to which the actual data are realized in our world. Broadly speaking, this is in line with the correspondence theory of truth similarly in \cite{Tarski:44}.

\begin{remark}
\label{rem:truepro}  
More detailed discussions on \cref{def:truepro}, including examples, are provided in Appendix D.
\end{remark}

Now that we have defined  \emph{learning} and \emph{true probability}, let us discuss under what conditions machines can or cannot learn the true probabilities. Before we move on, however, let us briefly mention how we can provide formal conditions for learning even though \cref{def:learn} contains informal notions. 

Recall from the second comment on \cref{def:effeccal} that the general notion of computation has not been mathematically defined. This is why the Church-Turing thesis remains as a thesis, not as a theorem, given that it uses the general notion of computation. But the computability of any target function in each \emph{specific} case can be formally specified by giving some definite and explicit instructions to derive the target function in each case, say by a Turing machine. Likewise, our general notion of machine learning cannot be mathematically defined because \cref{def:learn} uses the general notion of computation and the informal notion of success. But this does not prevent us from mathematically analyzing the notion of \textit{machine learning on the true probabilities} by proving what the necessary and sufficient conditions are to learn them. We can do so by giving some definite and explicit instructions to statistically derive the true probability function by a machine while satisfying the Success Criterion (1).   

\section{Kinds of Probabilities and Learning}

\subsection{Subjective vs. Objective Probabilities}

Broadly speaking, probabilities can be divided into two kinds, subjective and objective ones. Subjective probability, say $\Pi(A_{t+1}|$\ss $_{t})$, depends on each person's belief and thus possibly varies from person to person, while objective one, say $P(A_{t+1}|$\ss $_{t})$, does not. 

The standard theory of subjective probability was first developed by Ramsey and then further by De Finetti and Savage. Subjective probability is designed to represent a \textit{degree of belief} possessed by a subject, say some person or, if possible, a machine. Hence subjective probability represents whatever is in any one's mind upon anything as long as his/her belief system is coherent, and so can be assigned even to what is merely imagined. For example, while arguing for \textit{cogito, ergo sum,} \cite{Descartes:08} imagined an evil spirit that has devoted all its efforts to deceiving him. Descartes can assign some value of subjective probability to his imagination on the evil spirit in accordance with how likely it is to him that the imagination can be realized in this world, as long as Descartes' belief system remains coherent.

In contrast, objective probability, if any, is what must be determined by objective features of our world that do not vary from person to person. The best way to understand objective probability is to consider examples. Following \cite{Maher:10}, for example, suppose that a coin has the same face on both sides, that is, two-headed or two-tailed. When this coin is tossed infinitely often, its \textit{relative frequency} surely converges to 1 or 0. Hence the limiting relative frequency here is either 1 or 0, depending on how our world turns out to be, which is an objective matter, and not on whatever we believe. 

It should be noted that subjective and objective probabilities are conceptually bifurcated in two important ways. First, recall that subjective probability represents an aspect of someone's subjective belief, while objective probability does not. Hence the \textit{subjective} probability of Descartes' demon is positive as long as it is \textit{believed} at any degree that it could exist in our world. However, this does not necessarily imply that the \textit{true objective} probability of Descartes' demon is positive, since it might be the case that such a demon is possible only in one's imagination but impossible in our real world. We will return to this potential bifurcation between subjective and objective probability in Section 4.1.

Second, there exists an asymmetric relation between subjective and objective probability: although the subjective probability of Descartes' demon does not necessarily bind its objective probability, the converse holds. (e.g. \cite{Lewis:80}) That is, once it is \textit{proven/assumed} by any agent that the true objective probability of Descartes' demon is, say zero, then its subjective probability of the same agent is bound to this proven/assumed result on the objective probability and thus must be zero as well. From this asymmetric relationship, we derive \cref{lem:tolerror} in Section 4.2. 

\begin{remark}
\label{rem:probinterpretation}
More detailed discussions on various kinds of probabilities are provided in Appendix D.  
\end{remark}

\subsection{What is Implied by Learning the True Objective Probabilities?}

As we pointed out in Section 2, learning is the phenomenon of knowledge acquisition, and knowledge must be at least a true representation. In the case of human beings, the requirement of true representation is expressed as the requirement that (propositional) knowledge be at least a \textit{true belief} (e.g. \cite{Hintikka:62}, \cite{Moore:85}). What then is the counterpart of such a requirement for machines? 

In general, if a machine achieves computational success at $t$ by learning, what the machine represents by learning must be at least true at that time. Then we denote the true representation of the machine about what is learned by the ``true belief'' of the machine, a legitimate analogue to the true belief of human beings. It is a belief \textit{analogue}, for we haven't yet shown that machines have minds or that they have the same kinds of mental representations as human beings. It is nevertheless a \textit{legitimate} belief analogue, since the computational models of machine intelligence are based on understanding human intelligence. (e.g. \cite{Pearl:18}, \cite{Russell:98}, \cite{Valiant:84,Valiant:08})

That said, let us discuss the relation between belief and learning on the machine side: the knowledge acquired by machine learning must be at least a true belief. In \cite{Hintikka:62}, the knowledge of a person $i$ refers to the knowledge of that person $i$ on any proposition $A$. Likewise, machine's learning of the true objective probability $P$ here refers to the knowledge acquired by any machine on the probabilistic proposition $A_{p}$. If a machine learns the true probability as $\alpha$, then the probabilistic proposition $A_{p}$ amounts to that the true objective probability $P,$ if any, is what the very machine calculates as $\alpha$. Here, we convert the non-propositional learning into propositional learning.

Now, just as a person $i$'s knowledge on proposition $A$ must satisfy the necessary condition that the person $i$'s belief in $A$ is true, machine learning of the true probability $P$ must also satisfy the condition that the belief in $A_{p}$ of the machine is true. Note here that such a belief in $A_{p}$ is true when what has been calculated by the machine is indeed equal to the true probability $P$. Now, this calculated probability function by a machine is nothing more than the subjective probability of the machine. Therefore, the necessary condition for machine learning of true probability $P$ requires a machine to hold a true belief whose truth condition is satisfied when its subjective probability is, in fact, in congruence with the true objective probability $P$. In short, if a machine learns the true objective probability $P$, then the subjective probability $\Pi$ of the machine is actually equal to the true probability $P$.

 \begin{remark}
 \label{rem:literature}
There has been a large literature in logic and economics whose discussion implies when a machine holds a true belief in the probabilistic proposition $A_{p}$. We provide some literature in Appendix B.      
 \end{remark}

Therefore, we obtain the following condition:

\textbf{The Necessary Condition for any Machine to Learn the True Probability}
  
(2) If a machine \textit{learns} the true objective probability $P(A_{t+1}|$\ss $_{t})$, then $\Pi(A_{t+1}|$\ss $_{t})=P(A_{t+1}|$\ss $_{t})$ 

\ \ \ \ where
$\Pi(A_{t+1}|$\ss $_{t})$ denotes the subjective probability of the machine at time $t$. 

We assume, without loss of generality, that the event $A_{t+1}$ is an elementary event, for simplicity. So the event $A_{t+1}$ is a singleton, i.e. $\{\omega_{t+1}\}$.

Two things should be noted from (2): first, \textit{learning/knowledge} is not necessarily equivalent to \textit{obtaining true fact} that $\Pi(A_{t+1}|$\ss $_{t})=P(A_{t+1}|$\ss $_{t})$, as the converse of condition (2) does not necessarily hold. Second, if a machine is wrong in calculating the true probability at time $t$ so that $\Pi(A_{t+1}|$\ss $_{t})\neq P(A_{t+1}|$\ss $_{t})$, then by \textit{modus tollens} we can derive from (2) that the machine does not learn it at that time. However, this does not preclude the machine from learning it at any other time. Then what can be said about learnability in general? According to the Success Criterion (1), a machine \textit{cannot} learn any target function if it is wrong most of the time, except for a few finite cases out of infinite opportunities to learn. But can a machine be said to learn if it is correct infinitely often but also wrong as that often? We give a negative answer to this question by proving theorems in Section 4.2.

\section{Can Machines Learn the True Probabilities?}

\subsection{Learning the True Probabilities and Calibration}

Let us start with a simple example in which a machine is trying to learn the true probability that it will rain tomorrow. A forecasting system is said to be \textit{well-calibrated} if it assigns probability, say
30\%, to rainy events in a test set whose long-term proportion that actually rains is 30\%. According to \cite{Dawid:82}, a forecasting machine is \textit{self-assured} that its fairly arbitrary test set of forecasts is well-calibrated. This is \cref{thm:dawid}. In addition, we prove in \cref{thm:learncal} that if the machine \textit{learns} the true probability, then this machine's forecasting is \textit{truly guaranteed} to be well-calibrated.

Now, let us assume that a machine has its \emph{own} (not necessarily true in our context) probability distribution $\Pi$ defined over \ss $_{\infty}=
{\textstyle\bigvee\limits_{t=0}^{\infty}}\text{\ss }_{t},$ where \ss $_{t}$ is denoted by the totality of the true facts up to day $t$. The probability forecasts $\Pi(A_{t+1}|\text{\ss }_{t})$ it makes on day $t$ are for events $A_{t+1}$'s in \ss $_{t+1}$ and are \ss $_{t}$-measurable. For each day $t$ we have an arbitrary \textit{associated} event $A_{t}\in$ \ss $_{t}$, say the event of raining on day $t$. We denote the indicator of $A_{t+1}$ by $Y_{t+1}=1_{\{A_{t+1}\}}$, and introduce $\hat{Y}_{t+1}=\Pi(A_{t+1}|$\ss $_{t})$, the probabilistic forecast of machines on day $t$+1. In addition, we introduce the new indicator variables $\xi_1,\xi_2,\ldots,$ at choice to denote the inclusion of any particular day $t$ in the test set where $\xi_t=1$ if the day $t$ is included in the test set and $\xi_t=0$ otherwise. Now, if we set the selection criterion to include any day into the test set as the assessed probability $\alpha$ on day $t$, then we have the following theorem.

\begin{theorem}
\label{thm:dawid}
Suppose that $\xi_{t}\ $is \ss $_{t-1}$ measurable. Then, $\Pi $ $ (p_{k}\rightarrow$ $\alpha) =1$ when $k\to\infty$,

\ \ \ \ \ \ \ \ where \ \ \ \ \
$k$: the number of days in the test set

\ \ \ \ \ \ \ \ \ \ \ \ \ \ \ \ 
$p_k=(
{\textstyle\sum\limits_{t=1}^{k}}
\xi_t)^{-1}\cdot(
{\textstyle\sum\limits_{t=1}^{k}}
\xi_t\cdot 1_{\{A_{t+1}\}})$ 

\begin{center}
$\xi_{t}:=\begin{cases} 1 & \hat{Y}_{t+1}=\Pi(A_{t+1}|$\ss $_{t}) = \alpha \\ 0 & \hat{Y}_{t+1}=\Pi(A_{t+1}|$\ss $_{t}) \neq \alpha \end{cases}$ \end{center}
\end{theorem}

Here, let us use the terms as follows: machine forecasts are \textit{ self-assured} to be well-calibrated when $\Pi $ $ (p_{k}\rightarrow$ $\alpha) =1$, while those are \textit{truly guaranteed} to be so when $P $ $ (p_{k}\rightarrow$ $\alpha) =1$. It should be noted then that even if the forecasting machine is \textit{self-assured} to be well-calibrated, this does not necessarily imply that its forecasts are \textit{truly} \textit{guaranteed} to be well-calibrated. Recall from Section 3.1 that there is a conceptual bifurcation between subjective and objective probability. 

Now, suppose that a machine tries to learn the true probability of a particular event $A_{t+1}$. If this machine indeed \textit{learns} the true probability of the event as $\alpha$, then the machine should correctly calculate the true probability of the same events repeatedly as $\alpha$ most of the time. Hence, the machine can construct a test set of those associated events $A_{t+1}$'s whose sequentially \textit{correct} probabilities are $\alpha$. Then we can show further from \cref{thm:dawid} that the test set will be well-calibrated with \textit{true probability} $P$- one. This is \cref{thm:learncal}. In short, here ``being \textit{correct} as $\alpha$'' itself serves as what \cite{Dawid:82} calls a \textit{selection criterion}.

However, note that if the size of \ss $_{t}$ continues to grow as $t$ goes to infinity, then \ss $_{t}$'s might be different for each $t.$ Then $P(A_{t+1}|\text{\ss }_{t})$ might not stay the same as $\alpha$ even for the same events $A_{t+1}$'s across infinitely many $t$'s. Now, in order for the \textit{correct} probability $\alpha$ to work as a selection criterion, it should be that $P(A_{t+1}%
|\text{\ss }_{t})$ stays the same as $\alpha$ at least for infinitely many $t$'s even though \ss $_{t}$ may vary as time passes. Therefore, we prove \cref{lem:learnsel} from the following three assumptions. The justifications for the three assumptions are provided in Appendix C.

\begin{assumption}
\label{ass:assumption1}  
 \ss $_{t}$'s in $P(A_{t+1}|$\ss $_{t})$ are the set of all the \textit{true facts} up to time $t$.
\end{assumption}

\begin{assumption}
\label{ass:assumption2}  
 No further knowledge requirement is imposed on condition \ss $_{t}$.
\end{assumption}

\begin{assumption}
\label{ass:assumption3}  
Once a probability of an event type $E$ is established, its associated event tokens $E_{t_{k}}$'s \textit{occur} at some infinite subsequence of time $t_{k}$' s, so that $P(E_{t_{k}})$ does not vanish to zero as $t_{k}\rightarrow\infty$.
\end{assumption}

It should be noted from \cref{ass:assumption1} and \cref{ass:assumption2} that if \ss $_{t}$ is the set of \textit{known} facts, the information on the associated events $E_{t}$'s in {\ss}$_{t}$'s may not be independent of one another over time. Once $E_{t}$ has been known in the past at some time $t_{0},$ the same events $E_{t}$'s are more likely to be known afterwards. Repeatedly accumulated knowledge of the same events reinforces the probability that the very event will be known again in the future. However, this is not necessarily the case with the set of true facts. It will be clear in \cref{lem:learnsel} why this independence condition matters.

\begin{lemma}
\label{lem:learnsel}
For any $\alpha\in\Re\lbrack0,1]$, let $E_{t}$ denote the event token at time $t\in\mathbb{N}$ whose event type $E$ almost surely determines the true probability of an event type $A$ as $\alpha$. Then, if for some subsequence $t_{k}$'s, $E_{t_{k}}$'s are independent across $t_{k}$'s and $P(E_{t_{k}})\neq0$ for any $t_{k},$ then $P(E_{t}$ $i.o)=1.$
\end{lemma}

Now that \cref{lem:learnsel} has been established, $P(A_{t+1}|$\ss$_{t})$ is truly guaranteed to stay as $\alpha$ infinitely often, and thus the machine has infinite opportunities to learn $P(A_{t+1}|$\ss$_{t})$ as $\alpha$.

\begin{theorem}
\label{thm:learncal}
Let us consider any arbitrary $\alpha\in\Re\lbrack0,1]$. If a machine \textit{learns} the true objective probability $P(A_{t+1}|$\ss$_{t})$ as $\alpha$, then $P ($ $p_{k}\rightarrow$ $\alpha$ $) = 1$.
\end{theorem}
 
It should be noted that the notion of learning in \cref{thm:learncal} is flexible enough to allow for some finitely few potential errors, so that there can exist some $t^{\ast}<\infty$ such that $P(A_{t+1}|$\ss$_{t})\neq\alpha$ $\forall t < t^{\ast}$ while processing the data to learn. 

\begin{remark}
\label{rem:learncal}
More detailed discussions on \cref{thm:learncal} are provided in Appendix D.   
\end{remark}

\subsection{Can Machines Learn the True Probabilities?}

\begin{theorem}
\label{thm:impossdawid}
It is impossible to obtain a joint distribution for an infinite sequence of events that could have the well-calibration property with \textit{subjective} probability $1$.
\end{theorem}

The basic idea in the proof of \cref{thm:impossdawid} starts with constructing a \textit{counterexample} in which the true probability function $P$ is \textit{deviated} infinitely often from the subjective probability function $\Pi$ in such a way that the well-calibration property does not hold any longer.

\textbf{Counterexample 1} Following \cite{Oakes:85}, let $P$ be such as $P(A_{t}|$\ss $_{t-1})=f(\Pi
(A_{t}|\text{\ss }_{t-1})),$ with the function $f([0,1])\rightarrow
\lbrack0,1]$ being defined by $f(x)=x+\frac{1}{2}$ $(0\leq x\leq\frac{1}{2}),$
$f(x)=1-x$ $(\frac{1}{2}<x\leq1)$ for any event $A_{t}.$ Then, under $P$ with
$P(Y_{I_{k}}=1)=f(\alpha)$ where $\hat{Y}_{t}=$ $\alpha$ for a subsequence
$\{t:t=I_{1},I_{2},\ldots\}$ and $Y_{I_{k}}$'s form a Bernoulli sequence, the well-calibration property does not hold. 

Due to this counterexample from \cite{Oakes:85}, the machine forecaster cannot exclude the possibility that its
test set may be mis-calibrated, and thus the machine cannot hold its subjective probability $\Pi$- one of being well-calibrated. Furthermore, if this \textit{artificially-imagined} possibility of mis-calibration is a \textit{real} possibility, then it is derived that no test set large enough can be \textit{guaranteed} to be well-calibrated with the true probability $P$- one. Later in this section, we prove that if such an imagined possibility is a real one, then machines cannot learn. Meanwhile, we also prove mathematically how the \cite{Oakes:85} Counterexample paralyzes Dawid's \cref{thm:dawid}, which amounts to the proof of \cref{thm:impossdawid}. 

\begin{remark}
\label{rem:count1}
More detailed discussion on the Counterexample 1 is provided in Appendix D.    
\end{remark}

\begin{lemma}
\label{lem:lemma2}
Suppose that a machine constructs a test set by the assessed probability $\alpha$. Then $E$ $|p_{\infty}-$ $\alpha|$ $=$ $0$ if and only if $P(p_{k}\rightarrow$ $\alpha)=1$ where the expectation is taken with respect to the true probability $P$. Here, $p_{\infty}%
=\lim\limits_{k\rightarrow\infty}$ $p_{k}$.
\end{lemma}

\begin{lemma}
\label{lem:lemma3} 
Let us fix $\alpha\in\Re\lbrack0,1]$. Now, suppose that $p_{\infty}$ exists. Then $E$ $[p_{\infty
}-\alpha]=0$ if and only if $E[\lim\limits_{k\rightarrow\infty}\frac{1}{k}{\textstyle\sum\limits_{j=0}^{k-1}} P(A_{t_{j}+1}|\text{\ss }_{t_{j}})-\alpha]=0$. In general, $E$ $|p_{\infty}-\alpha|$ $\geq$ $E$ $|\lim\limits_{k\rightarrow\infty
}\frac{1}{k}{\textstyle\sum\limits_{j=0}^{k-1}} P(A_{t_{j}+1}|\text{\ss }_{t_{j}})-\alpha|.$
\end{lemma}

\begin{remark}
\label{rem:lemma23} 
By \cref{lem:lemma2} and \cref{lem:lemma3}, we establish a connection between the true guarantee of well-calibration and the \textit{real} forecasting \textit{game }between a machine and Nature. More discussions on such connection by \cref{lem:lemma2} and \cref{lem:lemma3} are provided in Appendix D.
\end{remark}

\begin{definition}
\label{def:perverse} 
Nature is \textit{perverse} when, for any fixed machine forecast $\alpha$, $P(A_{t+1}|$\ss $_{t})\neq\alpha$ \textbf{at least} for infinitely many $t$'s along the stochastic path of the \textit{test set}.
\end{definition}

By ``\textbf{at least} \textit{i.o.}'' in \cref{def:perverse}, we mean that Nature deviates from $\alpha$ either \textit{ (i) infinitely often} or \textit{(ii) all but finitely often} along the stochastic path of the test set. Thus, we clearly distinguish (\textit{i}) from (\textit{ii}). From now on, we mean by ``infinitely often'' that nature not only deviates infinitely often, but also does not deviate infinitely often. On the other hand, by ``all but finitely often'' we mean as usual. Then, if the true probability of Nature's perversity is zero, then we denote it by $P(P(A_{t+1}|${\ss}$_t)\neq\alpha$ \textbf{at least} $i.o.$ along the path of the test set$)=0$, which amounts to $P(P(A_{t+1}|${\ss}$_{t})\neq\alpha$ \textbf{at most} for $t < \infty$ along the path of the test set$)=1$. Furthermore, if there is no confusion, we will simplify Nature's perversity by ``$P(A_{t+1}|$\ss $_{t})\neq\alpha$ at least $i.o.$'' while omitting ``along the path of the test set.'' 

Now, according to the Success Criterion (1), a machine fails to learn the true probability in case \textit{(ii)}, because the machine then makes wrong forecasts along the path except for a finite few of the infinite opportunities to learn. However, it seems unclear whether the machine can learn or not in case \textit{(i)}. On the one hand, the machine seems not to be able to learn because it makes too many errors, say infinitely many errors. On the other hand, it seems that the machine should be able to learn because it makes astronomically many correct forecasts, say infinitely often. Therefore, while adopting this definition, we clearly prove by \cref{thm:cannotlearn1} and \cref{cor:corollary1} that a machine cannot learn the true probability even when it is correct infinitely often, if it is wrong that often.       

\textbf{Observation} Provided that the machine forecast $\Pi(A_{t+1}|$\ss $_{t})$ is fixed as some value $\alpha\in
\Re\lbrack0,1]$, $P($ $\Delta_{t}$ $)$ becomes the true second-order probability on the true first-order probability of such event $A_{t+1}$, that is, $P($ $\Delta_{t}$ $)$ = $P$
$\left(\text{ }P(A_{t+1}|\text{\ss }_{t})=\alpha\text{ }\right)$ where $\Delta_{t}$ denotes the event that the machine makes a correct forecast at $t$. 

It should be noted here that the computable numbers by a machine are countably many (e.g. \cite{Turing:36}). Thus, the true second-order probability $P$ here is a probability \textit{mass} function on countable space and therefore satisfies the Kolmogorov axioms, although $\alpha$ may potentially be any real number in $\Re\lbrack0,1].$ 

\begin{remark}
\label{rem:2ndprob} 
More detailed discussions on the connection between true second-order probability and the forecasting game are provided in Appendix D.
\end{remark}

\begin{lemma}
\label{lem:lemma2ndprob} 
Let us consider the forecasting game between Nature and a machine. Also, let us further suppose that the structure of this game at any given time $t$, i.e. whether it is simultaneous or not, is certain to Nature. Now, by \cref{ass:assumption1} and \cref{ass:assumption2}, let us suppose that \ss $_{t}$ consists of the true facts, not necessarily knowledge. Then there exists a true second-order probability $P$ such that $0 < P\left(  P(A_{t+1}|\text{\ss }_{t})=\alpha\right) < 1
$ if and only if the \textit{real} forecasting game is a
simultaneous-move game at time $t$. In particular, $P$
$\left(  P(A_{t+1}|\text{\ss }_{t})=\alpha\right)=0$ if and only if the machine moves first and then Nature moves later after observing what move the machine takes in the forecasting game at time $t$. 
\end{lemma}

There are various theories of learning in games. (e.g. \cite{Nisan:07}) Therefore, what matters is what is aimed to learn through games and who are competing with each other in the games. In the standard model, a machine aims to learn what the optimal actions are to produce the minimized expected (total) loss or payoff, which is determined in a given environment, say financial market. In this case, a machine usually competes with other machines in the game. For example, in some online learning, a machine aims to learn a sequence of estimates which return the sub-linear regret, given that the loss functions are convex. It gets a possibly different amount of payoff/loss at each round of games along the stochastic path where the given sequence of games are played. 

In our forecasting games, on the other hand, a machine aims to learn the true objective probability, if any, through games, and so the machine is competing with Nature in the game. Also, whoever wins a game, the winner/loser will get uniform payoff at every round along the path, for what counts is how many times the machine loses/wins along the path, not how much payoff it gets at each round along the path once it loses/wins. 

\begin{theorem}
\label{thm:winninglearning}
In the forecasting game between a machine and Nature, the machine does not necessarily learn that it wins at each round of the game even though it indeed wins.
\end{theorem}

Thus, winning strategy is not equivalent to learning strategy. Now, in case when a machine does not learn that it wins/loses a game even though it indeed does so, it does not matter what it gets as payoff when it wins/loses because it cannot learn how much it gets at each round. What matters, on the contrary, is how many times it wins along the path, and this is why our game setting in \cref{lem:lemma2ndprob} adopts a uniform payoff at each round.       
\begin{theorem}
\label{thm:3cases}
Let us consider any arbitrary $\alpha\in\Re\lbrack0,1]$ for any machine forecast. If $P(p_{k}\rightarrow$ $\alpha)=1$, then the true
probability that Nature is perverse is zero with any of these forecasts $\alpha$. $($\textbf{Case 3}$)$

$($\textbf{Case 1}$)$ Let us suppose that $P(A_{t+1}|$\ss$_{t})\neq\alpha$ at most finitely often along the stochastic path where the associated event $A_{t+1}$'s occur. Then $P(p_{k}\rightarrow$ $\alpha)=1$ where $p_{k}$ denotes the limiting relative frequency along the path.

$($\textbf{Case 2}$)$ Let us suppose that $P$$(P(A_{t+1}|$\ss$_{t})\neq\alpha$ just as in \cite{Oakes:85}$) \neq0$. Then, $P(p_{k}\rightarrow$ $\alpha) \neq1$ where $p_{k}$ denotes the limiting relative frequency along the stochastic path of the test set.

$($\textbf{Case 3}$)$ Let us suppose that $P$$(P(A_{t+1}|$\ss$_{t})\neq\alpha$ at least $\textit{i.o.}$ along the test set$) \neq0$. Then $P(p_{k}\rightarrow$ $\alpha)\neq1$ where $p_{k}$ denotes the limiting relative frequency along the path of the test set.
\end{theorem}

Regarding \cref{thm:3cases}, it is worth noting the following three things: (\textit{i}) $($\textbf{Case 1}$)$ is equivalent to the strong law of large numbers under a weaker assumption than \textbf{i.i.d.}: if the true probability $P(A_{t+1}|${\ss}$_{t})$ exists and $P(A_{t+1}|${\ss}$_{t})$ is identically distributed as $\alpha$ all but finitely often along the path, then the limiting relative frequency converges to the same $P(A_{t+1}|${\ss}$_{t})$ as $\alpha$ with true probability $P$- one. (\textit{ii}) $($\textbf{Case 2}$)$ shows that if \cite{Oakes:85} holds with $\Pi-$subjective probability $>0$, then \cite{Dawid:82} does not hold, which amounts to the proof of \cref{thm:impossdawid}. (\textit{iii}) $($\textbf{Case 3}$)$ shows, combined with \cref{thm:learncal}, that if $P$$(P(A_{t+1}|$\ss$_{t})\neq\alpha$ at most $f.o.$ along the test set$) \neq1$, then a machine \emph{cannot} learn the true probability $P(A_{t+1}|$\ss$_{t})$ as $\alpha$. Thus, the third result (\textit{iii}) has the following important implication for time-series analysis: a machine cannot relax the assumption that the true probability $P(A_{t+1}|$\ss$_{t})$ is identically distributed along the stochastic path, if the machine aims to learn the true probability $P(A_{t+1}|$\ss$_{t})$. To learn, the machine needs some identical distributional assumptions such as stationarity or ergodicity.

\begin{definition}
\label{def:uniperverse}
Suppose that, with true probability $P>0,$  Nature is perverse with some forecast $\alpha^{\ast}$. Then, Nature is \textit{uniformly perverse}, when for any forecast $\alpha\in\Re\lbrack0,1]$, there exists no $\alpha\neq\alpha^{\ast}$ such that $P($ $P(A_{t+1}|$\ss $_{t})\neq\alpha$ at least $i.o.) = 0$ for any event $A_{t+1}$.
\end{definition}

In other words, when Nature deviates from forecasters for any event $A_{t+1}$, she does not discriminate against some forecasters in favor of the others whose forecasts $\alpha$ Nature decides to conform to all but finitely often for sure.

\begin{theorem}
\label{thm:theorem5}
Suppose that, for any $\alpha$, there exists a true second-order probability $P$ such that $P (  P(A_{t+1}|\text{\ss }_{t})=\alpha\text{ })<1$ at least for infinitely many $t$'s. Then, Nature is uniformly perverse.
\end{theorem}

\begin{theorem}
\label{thm:cannotlearn1}
Suppose that, for any $\alpha$, there exists a true second-order probability $P$ such that $P (  P(A_{t+1}|\text{\ss }_{t})=\alpha\text{ })<1$ at least for infinitely many $t$'s. The machine cannot then learn the true objective probability $P(A_{t+1}|$\ss $_{t})$ as $\alpha$.
\end{theorem}

Now, let us discuss what it means in \cref{thm:cannotlearn1} by the condition that the true second-order probability is strictly less than 1. Note from \cref{lem:lemma2ndprob} that $P$ $\left(  P(A_{t+1}|\text{\ss }_{t})=\alpha\right)=1$ if and only if Nature moves first and then the machine moves later after observing what move Nature takes in the forecasting game at time $t$. Thus, it is clear from the condition of \cref{thm:cannotlearn1} why and when the machine fails to learn the true probability if Nature is uniformly perverse: when the machine cannot move later after observing the true move of Nature infinitely often, there always exists a real possibility that the machine may not be able to match Nature's move that often. Hence the machine cannot be truly guaranteed to be well-calibrated, which again implies the impossibility of machine learning. Since the machine cannot observe the true move of Nature in those forecasting games, the true probability is \textit{unobservable} by the machine. 

So far we have shown that it is of real possibility that Nature is perverse, and thus that no machines can learn the true objective probability. Now someone might argue that its proof holds only under the condition that Nature is uniformly perverse. Nature may not be uniformly perverse, however, but only selectively perverse, so that, for some forecast $\alpha_{0}$, Nature may decide to be benevolent enough to conform to that $\alpha_0$. Then it may be the case that the true probability of Nature being perverse is zero for this $\alpha_{0}$, and accordingly that machines may be given an opportunity to learn the true objective probability for that $\alpha_{0}$. 

Note, however, that it is entirely Nature's decision when she will be benevolent to a machine and when she will not. Therefore, it is still a random event to the machine whether Nature is perverse or not. If so, we will show further that, even if the true probability of Nature's being perverse is zero with some $\alpha_{0},$ a machine still cannot learn the true probability if it cannot learn which forecast is the right $\alpha_{0}$ for any event $A_{t+1}$. 

\begin{definition}
\label{def:tolerror}
A machine \textbf{tolerates error} at $t$ while pursuing its goal of learning the true probability $P(A_{t+1} | ${\ss}$_t)$, when $\Pi (A_{t+1} | ${\ss}$_t) = \alpha$ but $\Pi ( \{P(A_{t+1} | ${\ss}$_t ) \neq \alpha\}) > 0$ for some $\alpha\in\Re\lbrack0,1].$ 
\end{definition}

\begin{remark}
\label{rem:tolerror}
In relation to \cref{lem:tolerror}, more detailed interpretation on \cref{def:tolerror} is provided in Appendix D.
\end{remark}
 
\begin{lemma}
\label{lem:tolerror}
Suppose that a machine aims to learn the true probability $P(A_{t+1} | ${\ss}$_t)$ and thus performs an effective calculation to return its result of $\Pi (A_{t+1} | ${\ss}$_t)$ as $0$ for the true probability $P(A_{t+1} | ${\ss}$_t)$. Then, $\Pi (A_{t+1} | ${\ss}$_t) = 0$ if and only if $\Pi ( \{P(A_{t+1} | ${\ss}$_t ) = 0\}) = 1$, for all but finitely many $t$'s.
\end{lemma}

\begin{remark}
\label{rem:savage}
In relation to \cite{Savage:72}, more discussions on \cref{lem:tolerror} are provided in Appendix D.
\end{remark}

\begin{definition}
\label{def:selperverse}
Nature is \textit{selectively} perverse, when $\exists $ $ \alpha$ and $\alpha_{0}\neq\alpha$\ such that $P(P(A_{t+1}|$\ss$_{t})\neq\alpha_{0}$ at least $i.o.)=0\,$,$\ $while $P(P(A_{t+1}|$\ss$_{t})\neq\alpha$ at least $i.o.)>0$ for any other $\alpha\neq\alpha_{0}
$.
\end{definition}

Now, let us define Nature's decision to be selectively perverse at $t$ to show by \cref{lem:stoppingtime} that once Nature decides so at $t$, our real world remains as such. 

\begin{definition}
\label{def:decision}
Nature \textit{decides} to be selectively perverse at $t$, when there exist forecasts $\alpha$ and $\alpha_{0}\neq\alpha$\ such that $P(A_{\alpha_{0}}%
(t+1)|$\ss $_{t})=0$, while $P(A_{\alpha\neq\alpha_{0}}(t+1)|$\ss $_{t}%
)\neq0$ where $A_{\alpha}(t+1)$ denotes the event that, from $t+1$ onward, Nature is perverse with the associated events $A_{t}$'s whose assessed forecasts are $\alpha$. 
\end{definition}

\begin{definition}
\label{def:stoppingtime}
Suppose that Nature is selectively perverse so that she freely decides at any time whether to be perverse at any rate or not. Then, $t_s<\infty$ denotes a \textit{stopping time} if $t_s$ is the last time that Nature changes her mind into non-perversity so that, for any $\alpha_0$ with which Nature is not perverse with true probability $P$-one, $P(A_{\alpha_{0}}%
(t+1)|$\ss $_{t})=0$, $\forall t > t_s$. 
\end{definition}

Note that $t_s$ is \ss $_{t}$-measurable, because \ss $_{t}$ includes all the true facts up to $t$ and so whatever Nature decides at $t$, say the event $\{ P(A_{\alpha_{0}}%
(t+1)|$\ss $_{t})=0 \}$ belongs to the set of true facts, \ss $_{t}$. 

\begin{lemma}
\label{lem:stoppingtime}
Nature is selectively perverse if and only if there exists a stopping time $t_s$ for every forecast $\alpha_0$ with which Nature is not perverse with true probability $P$-one so that $P(A_{\alpha_{0}}(t+1)|$\ss $_{t})=0$ $ \forall t > t_s$, while there is no stopping time $t_s$ for any other $\alpha\neq\alpha_0$.
\end{lemma}

\begin{lemma}
\label{lem:learnselfassured}
Let us suppose that Nature is selectively perverse and that a machine learns which forecast is the right forecast $\alpha_0$ for any associated $A_t$'s with which Nature is not perverse with true probability $P$- one. The machine is then self-assured that the stopping time $t_s$ arrives for that $\alpha_0$. 
\end{lemma}

\begin{corollary}
\label{cor:corollary1}
Suppose that Nature is \textit{selectively} perverse so that, with true probability $P$-one, she is not perverse with some machine forecasts $\alpha_0$. Furthermore, suppose that the machine is not self-assured that the stopping time $t_s$ arrives for each of those $\alpha_0$'s. The machine cannot then learn the true objective probability $P(A_{t+1}|$\ss$_t)$ as $\alpha$.
\end{corollary}

Note that along the stochastic path considered in \cref{cor:corollary1}, $P(P(A_{t+1}%
|$\ss $_{t})\neq\alpha_{0}$ at least $i.o.)=0$ $\forall t > t_s$. Now, for this $\alpha_0$,  

(3) \ \ \ \ $\limsup\limits_{t\rightarrow\infty}$ $P ( P(A_{t+1} | ${\ss}$_t  ) \neq \alpha_0)\leq P ( P(A_{t+1} | ${\ss}$_t  ) \neq \alpha_0$ at least $i.o )=0$ 

Therefore, without loss of generality, letting $t^{\ast} \geq t_s$ with $t^{\ast} < \infty$,

(4) \ \ \ \ $P ( P(A_{t+1} | ${\ss}$_t  ) = \alpha_0) = 1,$ $\forall t > t^{\ast} \geq t_s$ with $t^{\ast} < \infty$. 

Now, (4) means by \cref{lem:lemma2ndprob} that the true probability is observable at any time $t > t^{\ast}$ along this path. Then why is the machine still unable to learn the true probability, even though the machine can move after observing what move Nature takes at the forecasting games all along that path after $t^{\ast}$? According to \cref{cor:corollary1}, this is because the machine cannot be self-assured whether the true probability will remain observable at any time after $t^{\ast}$+1 onward, even if the machine observes Nature's true move at time $t^{\ast}$+1. Let us show this by the following \cref{lem:condcor1}.

\begin{lemma}
\label{lem:condcor1}
Suppose that a machine is not self-assured of the stopping time $t_s$ for $\alpha_0$. The machine cannot then be self-assured whether the true probability will remain observable at any time after $t^{\ast}$+1 onward, even if the machine observes Nature's true move at time $t^{\ast}$+1.
\end{lemma}

From \cref{thm:cannotlearn1} and \cref{cor:corollary1}, we conclude that the impossibility of learning is derived under the assumption either that Nature is uniformly perverse or that Nature is selectively perverse but a machine is not self-assured of whether the stopping time arrives or not. What would then happen in the case where Nature is selectively perverse and a machine is self-assured of the stopping time $t_s$ when the $t_s$ indeed exists? We show in the following that a machine can learn the true probability in this case, and further that this is the only case in which a machine can learn it. 

\begin{theorem}
\label{thm:learnselfassured}
Suppose that a machine learns the true probability $P(A_{t+1} | ${\ss}$_t  )$ as $\alpha$. The machine is then self-assured that the stopping time $t_s$ arrives for $\alpha$,  while the machine is not self-assured that the stopping time $t_s$ arrives for $\alpha$ where such $t_s$ does not exist.
\end{theorem}

Let us now define when the true probability is \textit{directly observable} based on the notion of \textit{population}. The concept of population in \cref{def:direcobs} is mainly indebted to \cite{Mises:57, Mises:67}. Since the true probability is defined as the \textit{empirical} distribution of this population available to a machine, the probability is said to be \textit{directly observable} by the machine.

\begin{definition}
\label{def:population}
 Let us consider a set $S$ that consists of the sequence of events $A_{t+1}$'s, $\{A_{t+1}\}_{t=0}^{k-1}$ with $k$ potentially infinite. Then, the set $S$ is defined to be a \textit{population} with $k$ number of elements, when this set $S$ is assumed to have a certain attribute of interest, and so an indicator variable $1_{\{A_{t+1}\}}$ is assigned to each event $A_{t+1}$ where $1_{\{A_{t+1}\}}$ has a value 1 or 0 depending on whether the event $ A_{t+1}$ satisfies such an attribute or not, once the set $S$ is collected. Then, the empirical distribution of the population $S$ with respect to the given attribute is defined to be $\frac{1}{k}\sum\limits_{t=0}^{k-1} 1_{\{A_{t+1}\}}$. 
 \end{definition}

\begin{definition}
\label{def:direcobs}
A machine \textit{directly observes} $P$$(  A_{t+1}|${\ss }$_{t})$ from the \textit{population} $S$ at $t^{\ast}$ if the following two conditions are satisfied: (\textit{i}) a population $S$ is in principle \textit{available} to the machine. (\textit{ii}) The machine calculates the \textit{empirical distribution} of the population with respect to the given attribute, which is the true probability distribution of the event $A_{t+1}.$ 
\end{definition}

 Now, in case where the sequence $\{A_{t+1}\}_{t=0}^{k-1}$ is a time-series, \cref{def:direcobs} means that $\Pi$$(  A_{t^{\ast}+1}|${\ss }$_{t^{\ast}})= \frac{1}{k}\sum\limits_{t=0}^{k-1} 1_{\{A_{t+1}\}}=P$$(  A_{t^{\ast}+1}|${\ss }$_{t^{\ast}})$ with $k=t^{\ast}$. Thus, when $t^{\ast}$ goes to infinity, the directly observable true probability becomes the limiting relative frequency, the representative objective true probability.

\begin{theorem}
\label{thm:selfassureddo}
Suppose that a machine is self-assured of the stopping time $t_s$ when there exists $t_s$, but that the machine is not self-assured of the stopping time $t_s$ when no $t_s$ exists. The machine then \textit{directly observes} the true probability $P$$(  A_{t+1}|${\ss }$_{t})$ as $\alpha_0$. 
\end{theorem}

\begin{theorem}
\label{thm:dolearning}
A machine \textit{directly observes} the true probability $P$$( A_{t+1}|${\ss }$_{t})$ as $\alpha$ if and only if the machine learns the true probability $P$$(  A_{t+1}|${\ss }$_{t})$ as $\alpha$.
\end{theorem}

Two things should be noted from \cref{thm:dolearning}. First, whenever the true probability is not directly observable, a machine cannot learn the true probability. Now recall from \cref{def:effeccal} that the machine is an ideal one with no practical limits on computational resources such as time or storage spaces. Therefore, this implies that no \textit{real} machines, hindered by many practical limits in our world, can overcome this impossibility of learning either, whenever the true probability is not directly observable. Second, \cref{thm:dolearning} also says that the true probability is directly observable by a machine whenever it can learn the true probability. Once a machine learns the true probability and so it is successfully computable, then the next question is how complex it is to compute. Now that the true probability is directly observable, this makes it easier to deal with the complexity problem. (e.g. Sorting algorithm) Thus, \cref{thm:dolearning} directly connects the problem of computational solvability to the problem of complexity. 

Now, let us finish this section by adding one more claim that the Success Criterion (1) to compute the true probability is sufficient for learning it. 

\begin{corollary}
\label{cor:successcriterion}
If a machine calculates the true probability $P$$(  A_{t+1}|${\ss }$_{t})$ correctly most of the time, which is self-assured to the machine, then the machine can learn the true probability.
\end{corollary}

\section{Conclusion}
We have discussed so far when machines can learn the true probabilities and when they cannot. In summary:

\begin{itemize}
\item $\exists $ $\alpha^{\ast}$ such that $P($ Nature is perverse with $\alpha^{\ast}$ $)>0$ by \cref{thm:theorem5}.
\end{itemize}

Now that Nature is perverse at least with one forecast $\alpha^{\ast}$,

\begin{itemize}
\item (\textit{i}) Nature is uniformly perverse: machines \textbf{cannot} learn by \cref{thm:cannotlearn1}.

\item (\textit{ii}) Nature is selectively perverse: $\exists$ $ $ $t_s$ for each $\alpha_0$ such that $P($ Nature is perverse with $\alpha_0$ $)=0$ by \cref{lem:stoppingtime}.
\end{itemize}

Then under (\textit{ii}),

\begin{itemize}
\item (\textit{ii}-1) Machines are not self-assured of the $t_s$: machines \textbf{cannot} learn by \cref{cor:corollary1}.

\item (\textit{ii}-2) Machines are self-assured of the $t_s$: 
\end{itemize}

Then under (\textit{ii}-2),

\begin{itemize}
\item (\textit{ii}-2-1) $t_s$ actually does not arrive: machines \textbf{cannot} learn by \cref{thm:learnselfassured}.
\item (\textit{ii}-2-2) $t_s$ indeed arrives: machines \textbf{can} learn and this is the \textbf{only} case in which machines can learn by \cref{thm:selfassureddo} and \cref{thm:dolearning}.
\end{itemize}

Before we close this section, let us add a few remarks. First, we emphasize that in this paper we have focused on the notion of ``machine learning'' that is not just a technical terminology, understood as an identification of a target function, but also an \textit{epistemic} one, a counterpart to ``human learning.'' We focus on this epistemic notion of machine learning because we particularly mean by ``machines'' those artifacts that perform human-level \textit{intelligent} behaviors.

Second, note that we do not need to specify \textit{how} machines learn the true objective probabilities to prove the impossibility of machine learning on the true probabilities. Instead, we only need the necessary condition for any machine to learn the true objective probabilities if it learns them in \textit{any} way. Thanks to this flexibility about how to learn, we come to have a powerful and robust result: no matter what kind of learning method a machine uses, it cannot learn the true objective probabilities that are not directly observable. 

Lastly, let us emphasize again that our learning machine is an ideal device with no practical limits on time and storage space, etc. Therefore, the scope and limit of machine learning on true probabilities discussed in this paper are more fundamental than practical ones.  

% Acknowledgements should only appear in the accepted version.
\section*{Acknowledgements}

The author is grateful to Tyler Burge, Michael Christ, Philip Dawid, Joseph Halpern, Jinho Kang, Steven Matthews, Thomas Sargent, and Byeong-uk Yi for discussions that were helpful in various ways to develop this paper. In particular, Tyler helped me pay attention to the idea of converting a non-propositional structure to a propositional one while learning, and Joe helped me open my eyes to the possibility of machine learning on the true probabilities. I discussed every detail of this paper with Jinho so that I insisted that he should be listed as a co-author. Jinho refused on the ground that he did not make direct contributions to mathematical proofs, with which I disagree. But Jinho has been right most of the time when we disagreed, so I decided to agree. Lastly, the author is grateful to three anonymous reviewers and a meta-reviewer. Their reviews were helpful in improving this paper.

\section*{Impact Statement}

This paper presents work whose goal is to advance the field of 
Machine Learning. There are many potential societal consequences of our work, none of which we feel must be specifically highlighted here.

% In the unusual situation where you want a paper to appear in the
% references without citing it in the main text, use \nocite
%\nocite{langley00}

\bibliography{example_paper}

\begin{thebibliography}{34}
\providecommand{\natexlab}[1]{#1}
\providecommand{\url}[1]{\texttt{#1}}
\expandafter\ifx\csname urlstyle\endcsname\relax
  \providecommand{\doi}[1]{doi: #1}\else
  \providecommand{\doi}{doi: \begingroup \urlstyle{rm}\Url}\fi

\bibitem[Blume \& Easley(2006)Blume and Easley]{BlumeE:06}
Blume, L. and Easley, D.
\newblock If you're so smart, why aren't you rich? belief selection in complete and incomplete markets.
\newblock \emph{Econometrica}, Vol. 74:\penalty0 929--966, 2006.

\bibitem[Blume \& Easley(2008)Blume and Easley]{BlumeE:08}
Blume, L. and Easley, D.
\newblock Market selection and asset pricing.
\newblock \emph{The Handbook of Financial Markets: Dynamics and Evolution}, by T. Hens IV and K. Schenk-Hoppe (ed.), North-Holland:\penalty0 403--438, 2008.

\bibitem[Boolos et~al.(2002)Boolos, Burgess, and Jeffrey]{Boolos:02}
Boolos, G., Burgess, J., and Jeffrey, R.
\newblock \emph{Computability and Logic}.
\newblock Cambridge University Press, Cambridge, 2002.

\bibitem[Carnap(1963)]{Carnap:63}
Carnap, R.
\newblock \emph{Logical {F}oundations of {P}robability}.
\newblock The University of Chicago Press, 1963.

\bibitem[Church(1936)]{Church:36}
Church, A.
\newblock An unsolvable problem of elementary number theory.
\newblock \emph{American Journal of Mathematics}, Vol. 58\penalty0 (2):\penalty0 345--363, 1936.

\bibitem[Cogley \& Sargent(2008)Cogley and Sargent]{SargentC:08}
Cogley, T. and Sargent, T.
\newblock The market price of risk and the equity premium: A legacy of the great depression.
\newblock \emph{Journal of Monetary Economics}, Vol. 55:\penalty0 454--476, 2008.

\bibitem[Cogley \& Sargent(2009)Cogley and Sargent]{SargentC:09}
Cogley, T. and Sargent, T.
\newblock Diverse belief, survival and the market price of risk.
\newblock \emph{Economic Journal}, vol. 119:\penalty0 354--376, 2009.

\bibitem[Dawid(1982)]{Dawid:82}
Dawid, P.
\newblock The well-calibrated bayesian.
\newblock \emph{Journal of the American Statistical Association}, 77\penalty0 (379):\penalty0 604--613, 1982.

\bibitem[Descartes(2008)]{Descartes:08}
Descartes, R.
\newblock \emph{Meditations on {F}irst {P}hilosophy}.
\newblock Translated by Moriarty. M. Oxford University Press, 2008.

\bibitem[Foster \& Vohra(1993)Foster and Vohra]{FosterV:93}
Foster, D. and Vohra, R.
\newblock Asymptotic calibration.
\newblock \emph{Biometrika}, 85\penalty0 (2):\penalty0 379--390, 1993.

\bibitem[Gaifman(1986)]{Gaifman:86}
Gaifman, H.
\newblock A theory of higher order probabilities.
\newblock In \emph{TARK}, 1986.

\bibitem[Halpern(2016)]{Halpern:16}
Halpern, J.
\newblock \emph{Actual Causality}.
\newblock The MIT Press, Cambridge, MA, 2016.

\bibitem[Halpern \& Fagin(1994)Halpern and Fagin]{HalpernF:94}
Halpern, J. and Fagin, R.
\newblock Reasoning about knowledge and probability.
\newblock \emph{Journal of the Association for Computing Machinery}, Vol 41\penalty0 (2):\penalty0 340--367, 1994.

\bibitem[Hintikka(1962)]{Hintikka:62}
Hintikka, J.
\newblock \emph{Knowledge and {B}elief}.
\newblock Cornell University Press, Ithaca, 1962.

\bibitem[Kozen(1997)]{Kozen:97}
Kozen, D.
\newblock \emph{Automata and Computability}.
\newblock Springer, New York, 1997.

\bibitem[Lewis(1980)]{Lewis:80}
Lewis, D.
\newblock A subjectivist's guide to objective chance.
\newblock \emph{Studies in {I}nductive {L}ogic and {P}robability, {V}olume II}, by R. Jeffrey (ed.):\penalty0 263--293, 1980.

\bibitem[Maher(2010)]{Maher:10}
Maher, P.
\newblock Explication of inductive probability.
\newblock \emph{Journal of Philosophical Logic}, Vol. 39:\penalty0 593â€“616, 2010.

\bibitem[Moore(1985)]{Moore:85}
Moore, R.~C.
\newblock A formal theory of knowledge and action.
\newblock \emph{Formal Theories of the Commonsense World}, by J. Hobbs and R. C. Moore, (ed.). Ablex Publishing Corp:\penalty0 319--358, 1985.

\bibitem[Nagel(1939)]{Nagel:39}
Nagel, E.
\newblock Principles of the theory of probability.
\newblock \emph{Int. {E}ncycl. {U}nif. {S}c.}, Vol. I\penalty0 (No. 6), 1939.

\bibitem[Nilsson(1986)]{Nilsson:86}
Nilsson, N.
\newblock Probabilistic logic.
\newblock \emph{Artificial Intelligence}, 28:\penalty0 71--87, 1986.

\bibitem[Nilsson(2011)]{Nilsson:11}
Nilsson, N.
\newblock \emph{Artificial Intelligence: A New Synthesis}.
\newblock Morgan Kaufmann, California, 2011.

\bibitem[Nisan et~al.(2007)Nisan, Roughgarden, Tardos, and Vazirani]{Nisan:07}
Nisan, N., Roughgarden, T., Tardos, E., and Vazirani, V.~V.
\newblock \emph{Algorithmic Game Theory}.
\newblock Cambridge University Press, New York, 2007.

\bibitem[Oakes(1985)]{Oakes:85}
Oakes, D.
\newblock Self-calibrating priors do not exist.
\newblock \emph{Journal of the American Statistical Association}, 80\penalty0 (390):\penalty0 p. 339, 1985.

\bibitem[Pearl(2018)]{Pearl:18}
Pearl, J.
\newblock A personal journey into bayesian networks.
\newblock \emph{UCLA Cognitive Systems Laboratory, Technical Report (R-476)}, 2018.

\bibitem[Ramsey(1931)]{Ramsey:31}
Ramsey, F.
\newblock Truth and probability.
\newblock \emph{Studies in Subjective Probability}, by Henry Kyburg and Howard smokler (ed.). Krieger:\penalty0 25--52, 1931.

\bibitem[Russell(1998)]{Russell:98}
Russell, S.
\newblock Learning agents for uncertain environments.
\newblock \emph{Proceedings of the Eleventh Annual Conference on Computational Learning Theory}, pp.\  101--103, 1998.

\bibitem[Sandroni(2000)]{Sandroni:00}
Sandroni, A.
\newblock Do markets favor agents able to make accurate predictions?
\newblock \emph{Econometrica}, Vol. 68\penalty0 (6):\penalty0 1303--1341, 2000.

\bibitem[Savage(1972)]{Savage:72}
Savage, L.~J.
\newblock \emph{The Foundations of Statistics}.
\newblock Dover Publications, New York, 1972.

\bibitem[Tarski(1944)]{Tarski:44}
Tarski, A.
\newblock The semantic conception of truth: and the foundations of semantics.
\newblock \emph{Philosophy and Phenomenological Research}, Vol. 4\penalty0 (3):\penalty0 341--376, 1944.

\bibitem[Turing(1936)]{Turing:36}
Turing, A.
\newblock On computable numbers, with an application to the entscheidungsproblem.
\newblock \emph{Proceedings of the London Mathematical Society}, 42:\penalty0 230--265, 1936.

\bibitem[Valiant(1984)]{Valiant:84}
Valiant, L.
\newblock A theory of the learnable.
\newblock \emph{Communications of the ACM}, 27, Nov.:\penalty0 1134--1142, 1984.

\bibitem[Valiant(2008)]{Valiant:08}
Valiant, L.
\newblock Knowledge infusion: In pursuit of robustness in artificial intelligence.
\newblock \emph{Proc 28th Conference on Foundations of Software Technology and Theoretical Computer Science}, pp.\  415--422, 2008.

\bibitem[von Mises(1957)]{Mises:57}
von Mises, R.
\newblock \emph{Probability, {S}tatistics and {T}ruth}.
\newblock revised English edition, Macmillan, New York, 1957.

\bibitem[von Mises(1967)]{Mises:67}
von Mises, R.
\newblock \emph{Mathmatical Theory of Probability and Statistics}.
\newblock 2nd edition, Academic Press Inc., New York, 1967.

\end{thebibliography}

%\bibliographystyle{icml2024}

%%%%%%%%%%%%%%%%%%%%%%%%%%%%%%%%%%%%%%%%%%%%%%%%%%%%%%%%%%%%%%%%%%%%%%%%%%%%%%%
%%%%%%%%%%%%%%%%%%%%%%%%%%%%%%%%%%%%%%%%%%%%%%%%%%%%%%%%%%%%%%%%%%%%%%%%%%%%%%%
% APPENDIX
%%%%%%%%%%%%%%%%%%%%%%%%%%%%%%%%%%%%%%%%%%%%%%%%%%%%%%%%%%%%%%%%%%%%%%%%%%%%%%%
%%%%%%%%%%%%%%%%%%%%%%%%%%%%%%%%%%%%%%%%%%%%%%%%%%%%%%%%%%%%%%%%%%%%%%%%%%%%%%%
\newpage
\appendix
\onecolumn
\section{Proofs for Lemmas, Theorems and Corollaries}
\bigskip

\textbf{Proof of \cref{thm:dawid} }A proof of \cref{thm:dawid} is suggested in \cite{Dawid:82}. A simpler one is as follows: Let $X_{t}=(%
%TCIMACRO{\tsum \limits_{j=1}^{t}}%
%BeginExpansion
{\textstyle\sum\limits_{j=1}^{t}}
%EndExpansion
\xi_{j})^{-1}\cdot\xi_{t}(Y_{t}-\hat{Y}_{t}).$ Since $(%
%TCIMACRO{\tsum \limits_{j=1}^{t}}%
%BeginExpansion
{\textstyle\sum\limits_{j=1}^{t}}
%EndExpansion
\xi_{j})^{-1},\xi_{t}$ and $\hat{Y}_{t}$ are \ss $_{t-1}$-measurable, it
follows that $E(X_{t}|$\ss $_{t-1})=0$ where $E$ is taken with respect to $\Pi(\cdot|$\ss $_{t-1})$ and so that $%
%TCIMACRO{\tsum \limits_{t=1}^{k}}%
%BeginExpansion
{\textstyle\sum\limits_{t=1}^{k}}
%EndExpansion
X_{t}$ is a martingale adapted to \ss $_{k-1}$. Also, $E((%
%TCIMACRO{\tsum \limits_{t=1}^{k}}%
%BeginExpansion
{\textstyle\sum\limits_{t=1}^{k}}
%EndExpansion
X_{t})^{2})=%
%TCIMACRO{\tsum \limits_{t=1}^{k}}%
%BeginExpansion
{\textstyle\sum\limits_{t=1}^{k}}
%EndExpansion
E(X_{t}^{2})\leq\lambda\cdot E\{%
%TCIMACRO{\tsum \limits_{t=1}^{k}}%
%BeginExpansion
{\textstyle\sum\limits_{t=1}^{k}}
%EndExpansion
((%
%TCIMACRO{\tsum \limits_{j=1}^{t}}%
%BeginExpansion
{\textstyle\sum\limits_{j=1}^{t}}
%EndExpansion
\xi_{j})^{-1}\cdot\xi_{i})^{2}\}\leq\frac{\lambda\pi^{2}}{6},$ because $Y_{t}$ is an indicator variable and so $var(Y_{t}|$\ss $_{t-1})$ is uniformly bounded above by some $\lambda$ such that $0 \leq\lambda < \infty$. Then, by
the martingale convergence theorem, $%
%TCIMACRO{\tsum \limits_{t=1}^{k}}%
%BeginExpansion
{\textstyle\sum\limits_{t=1}^{k}}
%EndExpansion
X_{t}$ converges with $\Pi-$probability one, which implies from Kronecker's
lemma that, with $\Pi-$probability one, $p_{k}-\alpha=(%
%TCIMACRO{\tsum \limits_{t=1}^{k}}%
%BeginExpansion
{\textstyle\sum\limits_{t=1}^{k}}
%EndExpansion
\xi_{t})^{-1}\cdot%
%TCIMACRO{\tsum \limits_{t=1}^{k}}%
%BeginExpansion
{\textstyle\sum\limits_{t=1}^{k}}
%EndExpansion
\xi_{t}(Y_{t}-\hat{Y}_{t})\rightarrow0$ where $\hat{Y}_{t}=\alpha $  $ \forall t.$ $Q.E.D.$

\textbf{Proof of \cref{lem:learnsel} }Let $A_{t}$ be an event token at time $t$ and
$P(A|E)=\alpha$ be the true probability of event type $A$ conditional on event
type $E$ whose event tokens are denoted by $A_{t}$ and $E_{t}$, respectively.
Then, by the definition of $E$ with respect to $A$, $P(A_{t+1}|E_{t}\in$ \ss $_{t})=\alpha$ with true probability $P-$ one. Now, once
$P(A_{t+1}|E_{t}\in$ \ss $_{t})$ is learned as such at some $t_{0},$ then
$E_{t_{0}}$ must have happened at that time and so $P(E_{t_{0}})\neq0.$ Also,
by \cref{ass:assumption3}, consider a subsequence of $E_{t_{k}}$'s where $P(E_{t_{k}%
})\neq0$ for any $t_{k}>t_{0}.$ Then, for this subsequence, $P(E_{t_{0}%
}\&E_{t_{k}})\neq0$ for any $t_{k}>t_{0},$ because $E_{t_{k}}$'s are
independent of one another$.$ 

Here, $E_{t_{k}}$'s are independent for the following reason: recall that by definition, $P(A_{t_{k}+1}|E_{t_{k}}\in$ \ss $_{t_{k}})=\alpha$ with true probability $P-$ one. Then, note that \ss $_{t_{k}}$ includes the fact that $P(A_{t_{k-i}+1}|E_{t_{k-i}}\in$ \ss $_{t-i})=\alpha$ for some $i\geq 1.$ Now, without loss of generality, let $i=1.$ Thus, we obtain

\bigskip
(1) \ \ \ \ \ \ $P($ $P(A_{t_{k}+1}|$ $\{P(A_{t_{k-1}+1}|E_{t_{k-1}})=\alpha\}$ $\in$ \ss $_{t_{k}})=\alpha)=1$  \ \ \ \ \ \ \ \ \ \ \   
\bigskip

Now that $E_{t_{k}}$ and $E_{t_{k-1}}$ are all included in \ss $_{t_{k}}$ by (1), to show that $E_{t_{k}}$'s are independent, we need to prove that 

\bigskip
(2) \ \ \ \ \ \ $P($ $\{P(A_{t_{k}+1}| ${\ss}$_{t_{k}})=\alpha \}$ $|$ $\{P(A_{t_{k-1}+1}| ${\ss}$_{t_{k-1}})=\alpha \} ) = P($ $\{P(A_{t_{k}+1}| ${\ss}$_{t_{k}}) = \alpha \} )$ \ \ \ \ 

But (2) is satisfied because $P($ $\{P(A_{t_{k}+1}| ${\ss}$_{t_k}) = \alpha \} ) = 1 = P($ $\{P(A_{t_{k-1}+1}| ${\ss}$_{t_{k-1}}) = \alpha \} )$.
\bigskip

Now that $P(E_{t_{0}%
}\&E_{t_{k}})\neq0$, for any $t_{k}>t_{0}$ in this subsequence$,$ we can always find some small enough $\epsilon>0$ such that $P(E_{t_{k}})>\epsilon.$ Therefore, the probability of the element in this subsequence does not vanish to zero, which implies that $\lim
\limits_{s\rightarrow\infty}P(E_{t_{0}}\&E_{t_{s}})\neq0.$ Since
$\lim\limits_{s\rightarrow\infty}P(E_{t_{0}}\&E_{t_{s}})\neq0$, $%
%TCIMACRO{\tsum \limits_{s=0}^{\infty}}%
%BeginExpansion
{\textstyle\sum\limits_{s=1}^{\infty}}
%EndExpansion
P(E_{t_{0}}\&E_{t_{s}})=\infty.$ Then, by the second Borel-Cantelli lemma,
$P(E_{t_{0}}\&E_{t_{s}}$ $i.o.)=1$ for $s>0,$ which means $P(E_{t_{0}}%
\in\text{\ss }_{t_{0}}$ $\&$ $E_{t_{k}}$ $\in$ \ss $_{t_{k}}$ $i.o.)=1$ for $t_{k}>t_{0},$
the desired result. $Q.E.D.$

\textbf{Proof of \cref{thm:learncal}} Suppose that, for infinitely many $t$'s when $P(A_{t+1}|$\ss $_{t})$ stays the same as $\alpha$, machines learn this
$P(A_{t+1}|$\ss $_{t})$ as $\alpha$ at time $t$. Then, by the Success Criterion (1), $\Pi(A_{t_k+1}|$\ss $_{t_k}) = \alpha = P(A_{t_k+1}|$\ss $_{t_k})$ at least infinitely often out of those infinite opportunities at $t$'s to learn. (We prove in \cref{cor:successcriterion} what we mean exactly by ``most of the time.'' Here we tentatively mean ``at least $i.o.$'' by it because machines are otherwise wrong too often to learn given the Success Criterion (1).)  Thus we can construct a test set which consists of the subsequence of
$\Pi(A_{t_{k}+1}|$\ss $_{t_{k}})$ which is equal to $P(A_{t_{k}+1}%
|\text{\ss }_{t_{k}})$ for those infinitely many $t_{k}$'s$.$ Let $\xi
_{t_{k}+1}=1$ if and only if $\Pi(A_{t_{k}+1}|$\ss $_{t_{k}})=P(A_{t_{k}%
+1}|\text{\ss }_{t_{k}})=\alpha.$ Note that $\xi_{t_{k}+1}$ is \ss $_{t_{k}}%
-$measurable, because machine forecasting $\alpha$ occurs at time $t_{k}$. Then, by \cref{thm:dawid},
with true probability $P-$one, $p_{k}- $ $\alpha = (%
%TCIMACRO{\tsum \limits_{j=0}^{k-1}}%
%BeginExpansion
{\textstyle\sum\limits_{j=0}^{k-1}}
%EndExpansion
\xi_{t_{j}+1})^{-1}\cdot%
%TCIMACRO{\tsum \limits_{j=0}^{k-1}}%
%BeginExpansion
{\textstyle\sum\limits_{j=0}^{k-1}}
%EndExpansion
\xi_{t_{j}+1}(Y_{t_{j}+1}-\alpha)\rightarrow0$, as $k\rightarrow\infty$ where
$P$ is defined over \ss $_{\infty}=%
%TCIMACRO{\tbigvee \limits_{t_{k}=0}^{\infty}}%
%BeginExpansion
{\textstyle\bigvee\limits_{k=0}^{\infty}}
%EndExpansion
\text{\ss }_{t_{k}}$ and \ss $_{t_{k}}$ is denoted by the totality of
\textit{true} \textit{facts} up to day $t_{k}.$ $Q.E.D.$

\textbf{Proof of \cref{lem:lemma2}} Clearly, if with $P-$probability one$,$
$p_{k}\rightarrow$ $\alpha,$ then $E$ $[p_{\infty}-$ $\alpha]$ $=0$ where the
mathematical expectation is taken with respect to the true probability $P,$
but not vice versa. The reverse does not necessarily hold, because even though $P($
$p_{k}\rightarrow$ $\alpha)<1,$ $E$ $[p_{\infty}-$ $\alpha]$ $=0$ when
$[p_{k}-\alpha]$ converges to $\pm\beta\neq0$ with the equal probability as
$\frac{1}{2}(1-P)>0.$ However, with $P-$probability one$,$ $p_{k}\rightarrow$
$\alpha$ if and only if $E$ $|p_{\infty}-$ $\alpha|$ $=0,$ for the following
reason: letting $\Lambda_{\infty}$ denote the event that $p_{k}\rightarrow$
$\alpha$ as $k$ goes to infinity, $E$ $|p_{\infty}-$ $\alpha|$ $=$
$P(\Lambda_{\infty})\times|p_{\infty}-$ $\alpha|_{\Lambda_{\infty}^{+}}$ $+$
$(1-P(\Lambda_{\infty}))\times|p_{\infty}-$ $\alpha|_{\Lambda_{\infty}^{-}}=0$
if and only if $P(p_{k}\rightarrow$ $\alpha)=1$ where $|p_{\infty}-$
$\alpha|_{\Lambda_{\infty}^{+}}$ denotes the value of $|p_{\infty}-$ $\alpha|$
when $\Lambda_{\infty}$ occurs, while $|p_{\infty}-$ $\alpha|_{\Lambda
_{\infty}^{-}}$ denotes that when $\Lambda_{\infty}$ does not occur. Here, the ``if" part is clear. For the ``only if" part, if $P(p_{k}\rightarrow$
$\alpha)<1,$ then $(1-P(\Lambda_{\infty}))\times|p_{\infty}-$ $\alpha
|_{\Lambda_{\infty}^{-}}>0$ while $P(\Lambda_{\infty})\times|p_{\infty}-$
$\alpha|_{\Lambda_{\infty}^{+}}=0$, which implies that $E$ $|p_{\infty}-$
$\alpha|$ $\neq0.$ $Q.E.D.$

\textbf{Proof of \cref{lem:lemma3}} By Fatou's lemma, $E[\liminf\limits_{k\rightarrow\infty}\frac{1}{k}{\textstyle\sum\limits_{j=0}^{k-1}}Y_{t_{j}+1}|$\ss$_{t_{j}}]$  $\leq $ $\liminf\limits_{k\rightarrow\infty}E[\frac{1}{k}
{\textstyle\sum\limits_{j=0}^{k-1}} Y_{t_{j}+1}|$\ss$_{t_{j}}] $ $= \liminf\limits_{k\rightarrow\infty}$ $\frac{1}{k}
{\textstyle\sum\limits_{j=0}^{k-1}} P(A_{t_{j}+1}|$\ss$_{t_{j}}) $ $\leq\limsup\limits_{k\rightarrow\infty}\frac{1}{k}{\textstyle\sum\limits_{j=0}^{k-1}} P(A_{t_{j}+1}|$\ss$_{t_{j}})\leq E[\limsup\limits_{k\rightarrow\infty}\frac{1}{k}{\textstyle\sum\limits_{j=0}^{k-1}} Y_{t_{j}+1}|$\ss $_{t_{j}}].$ Now, since $p_{\infty}$ exists by the assumption, $\liminf\limits_{k\rightarrow\infty}\frac{1}{k}
{\textstyle\sum\limits_{j=0}^{k-1}} Y_{t_{j}+1}=\limsup\limits_{k\rightarrow\infty}\frac{1}{k}
{\textstyle\sum\limits_{j=0}^{k-1}} Y_{t_{j}+1}.$ Then, by squeezing theorem, $\lim\limits_{k\rightarrow\infty
}\frac{1}{k}{\textstyle\sum\limits_{j=0}^{k-1}} P(A_{t_{j}+1}|$\ss $_{t_{j}})$ also exists and thus $E$ $[\lim
\limits_{k\rightarrow\infty}\frac{1}{k}
{\textstyle\sum\limits_{j=0}^{k-1}} Y_{t_{j}+1}$ $|\text{\ss }_{t_{j}}]=\lim\limits_{k\rightarrow\infty}\frac
{1}{k}{\textstyle\sum\limits_{j=0}^{k-1}} P(A_{t_{j}+1}|$\ss $_{t_{j}})$. Now, by the law of iterated expectations,
$E[\lim\limits_{k\rightarrow\infty}\frac{1}{k}
{\textstyle\sum\limits_{j=0}^{k-1}} Y_{t_{j}+1}]-\alpha=E$ $[E$ $[\lim\limits_{k\rightarrow\infty}\frac{1}{k}{\textstyle\sum\limits_{j=0}^{k-1}} Y_{t_{j}+1}$ $|\text{\ss }_{t_{j}}]-\alpha]=E[\lim\limits_{k\rightarrow\infty
}\frac{1}{k}{\textstyle\sum\limits_{t_{j}=0}^{k-1}} P(A_{t_{j}+1}|\text{\ss }_{t_{j}})-\alpha]$. Therefore, $E$ $[p_{\infty
}-\alpha]=0$ if and only if $E[\lim\limits_{k\rightarrow\infty}\frac{1}{k}{\textstyle\sum\limits_{t_{j}=0}^{k-1}} P(A_{t_{j}+1}|\text{\ss }_{t_{j}})-\alpha]=0.$ Also,$~E$ $|p_{\infty}-$
$\alpha|$ $=E$ $|\lim\limits_{k\rightarrow\infty}\frac{1}{k}{\textstyle\sum\limits_{j=0}^{k-1}} Y_{t_{j}+1}-\alpha|$ $=E$ $[E$ $[|\lim\limits_{k\rightarrow\infty}\frac{1}{k}{\textstyle\sum\limits_{j=0}^{k-1}} Y_{t_{j}+1}-\alpha|$ $|\text{\ss }_{t_{j}}]]$. But note that $E$ $[E$
$[|\lim\limits_{k\rightarrow\infty}\frac{1}{k}{\textstyle\sum\limits_{j=0}^{k-1}} Y_{t_{j}+1}-\alpha|$ $|\text{\ss }_{t_{j}}]]\geq E$ $|E[\lim
\limits_{k\rightarrow\infty}\frac{1}{k}{\textstyle\sum\limits_{j=0}^{k-1}} Y_{t_{j}+1}-\alpha|\text{\ss }_{t_{j}}]|$ $=E$ $|\lim\limits_{k\rightarrow
\infty}\frac{1}{k}{\textstyle\sum\limits_{j=0}^{k-1}} P(A_{t_{j}+1}|\text{\ss }_{t_{j}})-\alpha|$ by Jensen's inequality. Therefore, $E$ $|p_{\infty
}-\alpha|$ $\geq$ $E$ $|\lim\limits_{k\rightarrow\infty}\frac{1}{k}{\textstyle\sum\limits_{j=0}^{k-1}} P(A_{t_{j}+1}|\text{\ss }_{t_{j}})-\alpha|.$ $Q.E.D.$

\textbf{Proof of \cref{lem:lemma2ndprob}} Consider a simple two-player game $(I,S_{i},u_{i}(s))$ between Nature (player $i$) and a representative machine (player $-i$) where $I$ is the set of players $\{i,-i\}$,
$S_{i}$ is the set of pure strategies $s_{i}$'s for each player $i$, and $u_{i}(s)$ is the usual payoff function for player $i$. Since this is a probabilistic forecasting game, the pure strategy for each player $s_{i}$ can be any number in $\Re\lbrack0,1]$. But since the computable numbers by player $-i$ are countably many, we restrict $S_{i}$ to be countable. For simplicity, let $u_{i}:$ $S_{i}\times
S_{-i}\rightarrow\{-1,1\}$. In other words, for each profile $s=(s_{i}%
,s_{-i})$, if player $i$ wins, she obtains $1$, while she obtains $-1$
otherwise. When Nature (player $i$) succeeds in deviating from the machine forecast, Nature
wins. Otherwise, the machine (player $-i$) wins. Thus, this is a kind of matching game with countably infinite state space.

First, let us note that the structure of the forecasting game is given to Nature, because the structure
itself is something objective about the world and thus it belongs to the realm of Nature
herself. In other words, it is certain to Nature whether Nature and the machine moves simultaneously or not in the game as follows: If the machine moves when Nature herself does not move yet, then it is certain to Nature that the machine moves first and thus that it is not a simultaneous game. If the machine does not move yet when Nature does not move either, then it is certain to Nature that the machine does not move first, and thus whether it is a simultaneous game or not depends on Nature herself. If Nature reveals herself to the machine even before the machine moves so that the machine can move after observing Nature's, it is certain to Nature that it is not a simultaneous game. Otherwise, it is certain to Nature that it is a simultaneous game.

$(i)$ the proof of the ``only if'' part: first, let us fix machine forecast $\Pi(A_{t+1}|$\ss $_{t})$ as $\alpha$ and then consider the relevant test set. Now, suppose that the forecasting game along the stochastic path of this test set is not a simultaneous-move game at time $t$. Then, either Nature or the machine moves first, and the rest moves later after observing what move the other opponent takes. Thus, the one who can observe the opponent's move can control their/her own forecasting to win the game, and so $\Delta_{t}$ occurs or does not occur at time $t$, which is certain to Nature because the structure of the game is given to Nature. Then, since \ss $_{t}$ includes $\Delta_{t}$ or $\lnot\Delta_{t}$ as part of the true facts by \cref{ass:assumption1}, $P$ $\left(\Delta_{t}\in\text{\ss }_{t}\right)=1$ or $P$ $\left(\lnot\Delta_{t}\in\text{\ss }_{t}\right)=1$. Thus, it is either $P($ $P$ $\left(  A_{t+1}\text{}|\text{ }\Delta_{t}\in\text{\ss }_{t}\right)  =\alpha$ $) = 1$ or
$P($ $P$ $\left(  A_{t+1}\text{ }|\text{ }\lnot\Delta_{t}\in\text{\ss }%
_{t}\right)  =\alpha$ $)=0$ respectively, according as Nature moves first or the machine moves first. Therefore, the true second-order probability $P$ is neither strictly less than $1$ nor strictly greater than $0$.

$(ii)$ the proof of the ``if'' part: again, let us fix the machine forecast $\Pi(A_{t+1}|$\ss $_{t})$ as $\alpha$ and then consider the relevant test set. Now, suppose that the forecasting game is a simultaneous-move
game at time $t$. Then, for any fixed value $\alpha\in\Re\lbrack0,1]$, it is not certain to Nature herself
whether $\Pi(A_{t+1}|$\ss $_{t})=\alpha$ or not, because there exists no pure strategy Nash equilibrium in this simultaneous matching game. Thus, Nature cannot certainly control $P(A_{t+1}|$\ss $_{t})$ to make it deviate from $\Pi(A_{t+1}|$\ss $_{t})$ and so we obtain

\bigskip
(3) \ \ \ \ \ $P($ $P$$\left(  A_{t+1}\text{
}|\text{\ss }_{t}\right)  =\alpha$ $)\neq0$. 

\bigskip
(3) holds even though {\ss}$_t$ of $P$$\left(  A_{t+1}\text{
}|\text{\ss }_{t}\right)$ in (3) includes $\Delta_{t}$ or $\lnot\Delta_{t}$ as part of the true facts by \cref{ass:assumption1}, if either of them indeed occurs at $t$. In the same logic, it is not certain to Nature that the machine can control $\Pi(A_{t+1}|$\ss $_{t})$ to make it coincide with $P(A_{t+1}|$\ss $_{t})$ and so we obtain

\bigskip
(4) \ \ \ \ \ \ $P($ $P$$\left(  A_{t+1}\text{
}|\text{\ss }_{t}\right)  =\alpha$ $)\neq1$.

\bigskip
Clearly, any mixed strategy Nash equilibrium, if any, will lead to $0<P(P(A_{t+1}%
|$\ss $_{t})=\alpha$ $)<1$. Therefore, there exists the true second-order probability $P$ such that $0<P(P(A_{t+1}%
|$\ss $_{t})=\alpha$ $)<1$.

Furthermore, if Nature moves first, then $P($ $P$ $\left(  A_{t+1}\text{
}|\text{ }\text{\ss }_{t}\right)  =\alpha$ $)=1$, as we proved in $(i)$. Therefore, if the machine does not move first, which amounts to either Nature moves first or the machine moves simultaneously with Nature, then clearly $P($ $P$ $\left(  A_{t+1}\text{
}|\text{ }\text{\ss }_{t}\right)  =\alpha$$) \neq0$. $Q.E.D.$

\textbf{Proof of \cref{thm:winninglearning}} Consider the necessary condition (2) that if a machine \textit{learns} the true objective probability $P(A_{t+1}|$\ss $_{t})$, then $\Pi(A_{t+1}|$\ss $_{t})=P(A_{t+1}|$\ss $_{t})$. Since this is just a necessary but not sufficient condition, the converse of (2) does not necessarily hold. Now, for any machine forecast $\alpha\in \mathbb{R} [0,1]$, suppose that $P(A_{t+1}|${\ss} $_{t}) \neq \alpha $ for infinitely many $t$'s along the stochastic path where the associated $A_{t+1}$'s occur but that $P(A_{t+1}|${\ss} $_{t}) = \alpha $ for infinitely many $t^{\ast}$'s. Then, by \cref{thm:theorem5}, $P(P(A_{t+1}|$\ss$_{t})\neq\alpha$ $i.o. )>0$ for some event {$A_{t+1}$}. Thus, by (\textbf{Case 3}) of \cref{thm:3cases} and \cref{thm:learncal}, the machine cannot learn the true probability $P(A_{t+1}|$\ss $_{t})$, even though $\Pi(A_{t+1}|$\ss $_{t})=\alpha = P(A_{t+1}|$\ss $_{t})$ at infinitely many $t^{\ast}$'s. Thus, the machine does not learn that it wins even though it indeed wins at $t^{\ast}$'s. Clearly, the machine does not learn whether it wins at other $t$'s than $t^{\ast}$'s when it loses. Now, since the machine does not learn whether it wins or not at each round of game, the machine does not learn what its payoff is at each round. Furthermore, the machine is truly guaranteed to be well-calibrated along the path of $t^{\ast}$'s and so this is the winning strategy in forecasting game between Nature and the machine (e.g. \cite{FosterV:93}), but the machine still cannot learn the true probability $P(A_{t+1}|$\ss $_{t})$. Thus, in this case, winning strategy is not equivalent to learning strategy. $Q.E.D.$      

\textbf{Proof of \cref{thm:3cases}} First, let us recall the followings: by Nature's perversity with true probability $0$, we mean that
$P$( $M_{t}$ at least $i.o.)=0$ for any fixed $\alpha\in\Re\lbrack0,1]$. Here,
$M_{t}$ denotes a meta-event $\{P(A_{t+1}|$\ss $_{t})\neq\alpha$ for any event
$A_{t+1}$ at time $t\}$ for such a fixed forecast $\alpha$. Given this, let us
consider the following three cases, according as how
$P(A_{t+1}|$\ss $_{t})$ \textit{actually} varies with respect to $\alpha$
along the path of the test set. $($\textbf{Case 3}$)$ amounts to \cref{thm:3cases}.

\bigskip
$($\textbf{Case 1}$)$ Let us suppose that $P(A_{t+1}|$\ss $_{t})\neq\alpha$ for finitely
many $t$'s along the stochastic path. Now, as in \cref{thm:dawid}, let $X_{t}=(%
%TCIMACRO{\tsum \limits_{j=1}^{t}}%
%BeginExpansion
{\textstyle\sum\limits_{j=1}^{t}}
%EndExpansion
\xi_{j})^{-1}\cdot\xi_{t}(Y_{t}-\alpha).$ But, unlike in \cref{thm:dawid}, $\xi_{j}=1$ here if $P(A_{j+1}|$\ss $_{j})=\alpha$ for all $j$ along the stochastic path, not necessarily restricted to the test set. Now, consider those finite $t$'s when $P(A_{t+1}|$\ss $_{t})\neq\alpha$ and denote the largest $t$ among them by $t_{m}$. Then, $P(A_{t+1}|$%
\ss $_{t})-\alpha=E[Y_{t}|$\ss $_{t-1}]-\alpha=0$, $\forall t > t_{m}$ along the stochastic path. Thus, $E(X_{t}|$\ss $_{t-1})=0$ where expectation $E$ is taken with respect to the true probability $P(\cdot|$\ss $_{t-1})$ and so $%
%TCIMACRO{\tsum \limits_{t=t_{m+1}}^{k}}%
%BeginExpansion
{\textstyle\sum\limits_{t=t_{m+1}}^{k}}
%EndExpansion
X_{t}$ is a martingale adapted to \ss $_{k-1}$ at $t > t_{m}$ along the path. Then, by the martingale convergence theorem and Kronecker's lemma, $(%
%TCIMACRO{\tsum \limits_{j=0}^{k-1}}%
%BeginExpansion
{\textstyle\sum\limits_{j=0}^{k-1}}
%EndExpansion
\xi_{t_{j}+1})^{-1}\cdot%
%TCIMACRO{\tsum \limits_{j=0}^{k-1}}%
%BeginExpansion
{\textstyle\sum\limits_{j=0}^{k-1}}
%EndExpansion
\xi_{t_{j}+1}(Y_{t_{j}+1}-\alpha)\rightarrow0$ with true probability $P-$one.  

\bigskip
$($\textbf{Case 2}$)$ Let us consider the case where with true probability $P$ $>0$, $P(A_{t+1}|$\ss $_{t})$ deviates from $\alpha$ in such a way as
in Oakes (1985) along the test set. Then, $E$ $|p_{\infty}-$ $\alpha|$ $\neq0$
and so the calibration property is not truly guaranteed for the following
reason: Let $\Lambda_{\infty}^{o}$ be the event that $P(A_{t+1}|$\ss $_{t})$
deviates from $\alpha$ in such a way as in Oakes (1985) along the
test set. Then, since some subsequence of $Y_{t}$'s along the test set forms
Bernoulli whose relative frequency converges to $f(\alpha)\neq\alpha$, $p_{k}$
does not converge to $\alpha$ when $\Lambda_{\infty}^{o}$ occurs. Now, let
$|p_{\infty}-$ $\alpha|_{\Lambda_{\infty}^{o}}^{+}$ be the value of
$|p_{\infty}-$ $\alpha|$ when $\Lambda_{\infty}^{o}$ occurs, while
$|p_{\infty}-$ $\alpha|_{\Lambda_{\infty}^{o}}^{-}$ be the value of
$|p_{\infty}-$ $\alpha|$ when $\Lambda_{\infty}^{o}$ does not occur along the
test set. Then, in the same logic as in \cref{lem:lemma3}, we obtain that $E$
$|p_{\infty}-$ $\alpha|$ $=P(\Lambda_{\infty}^{o})\times|p_{\infty}-$
$\alpha|_{\Lambda_{\infty}^{o}}^{+}$ $+$ $(1-P(\Lambda_{\infty}^{o}%
))\times|p_{\infty}-$ $\alpha|_{\Lambda_{\infty}^{o}}^{-}$ $\neq0$. Thus,
$P(p_{k}\rightarrow\alpha)\neq1.$ However, the converse does not hold, for there can be many other ways of how $p_{k}$ does not converge to $\alpha$ than
in Oakes (1985). Hence it does not follow that $P(\Lambda_{\infty}^{o})>0,$
even if $E$ $|p_{\infty}-$ $\alpha|$ $\neq0.$ 

Now, suppose that with $\Pi-$subjective probability $>0,$ $P(A_{t+1}|$\ss $_{t})$ behaves in such a way as in Oakes (1985)$.$ Then, again in the same logic as in \cref{lem:lemma3}, we obtain that $E$
$|p_{\infty}-$ $\alpha|$ $=\Pi(\Lambda_{\infty}^{o})\times|p_{\infty}-$
$\alpha|_{\Lambda_{\infty}^{o}}^{+}$ $+$ $(1-\Pi(\Lambda_{\infty}^{o}%
))\times|p_{\infty}-$ $\alpha|_{\Lambda_{\infty}^{o}}^{-}$ $\neq0$ where expectation is now taken with respect to $\Pi$. Hence $\Pi(p_{k}\rightarrow\alpha)\neq1.$ Therefore, we conclude that if Oakes (1985) holds with $\Pi-$subjective probability $>0$, then Dawid (1982) does not hold, which amounts to the proof for \cref{thm:impossdawid}.

\bigskip
$($\textbf{Case 3}$)$ In general, suppose that the true probability of Nature's being perverse is not zero for any fixed forecast $\alpha$ on any associated events $A_{t}$'s. In other words, suppose that $P($ $M_{t}$ at least $i.o.$ along the test set$) > 0$ where $M_t$ is the meta-event that $P(A_{t+1}|$\ss$_{t})\neq\alpha$. Then, we claim that this implies that $E$ $|p_{\infty}-$ $\alpha|$ $\neq0$ where $E$ is taken with respect to $P$.

First, suppose that $p_{\infty}$ exists. Also, suppose that $\alpha\neq0$, because \textbf{(Case 3)} trivially holds if $\alpha = 0$. Now let us consider an infinite subsequence of $A_{t_{k}}%
$'s, $\{A_{t_{k_{j}}}\}_{j=0}^{\infty},$ which is conditionally identically distributed along the test set where $M_{t}$ occurs at least infinitely often. We can do this by Kolmogorov axioms 1 and 2 and \cref{lem:learnsel} for the following reason: note that by
Kolmogorov axioms 1 and 2 there always exists one $\beta\in\Re\lbrack0,1]$ such that $P(A|E)=\beta$ for any type event $A$ and $E,$ given that there exists probability of type event, if any. Then, for this $\beta$, $P($
$P(A_{t+1}|$\ss $_{t})=\beta$ $\ i.o.)=1$ according to \cref{lem:learnsel}. Thus, we found
one subsequence of $\{A_{t_{k}}\}_{k=0}^{\infty}$ such that it is
conditionally identically distributed as $\{P(A_{t_{k}+1}|$\ss $_{t_{k}%
})=\beta\}_{k=0}^{\infty}$. Now, fix $\alpha$. Also, without loss of generality, suppose that $\beta\neq\alpha$. Since $\beta
\neq\alpha$ is arbitrary, from this subsequence we can consider another
subsequence $E_A$ of $\{A_{t_{k_{j}}}\}_{j=0}^{\infty}$ with the true probability $P>0$
such that $E_A$ = $\{P(A_{t_{k_{j}}+1}|$\ss $_{t_{k_{j}}})=\beta\}_{j=0}^{\infty}$ along the stochastic path of the test set in which $M_{t}$ occurs at least infinitely often. 

For \textit{reductio}, let us suppose that Nature deviates $\alpha$ by picking numbers from uncountably many values of $\beta$'s such that every value of $\beta$ is equal to $P(A_{t+1}|$\ss $_{t})$ only at most finitely many $t$'s along the test set with true probability $P$- one. In other words,
\bigskip

\ \ \ \ \ (5) For $\beta\in\Re\lbrack0,1]$ where $\beta\neq\alpha$, $P(A_{t+1}|$\ss $_{t}) = \beta$ at most for finitely many $t$'s along the path of the test set where $M_t$ occurs at least infinitely often, with true probability $P$- one.
\bigskip

Note that there must be countably infinite number of different $\beta$'s in (5). Let us denote each different $\beta$ at each time along the path by $\beta_{t_{k_{j}}}$, while letting $\beta_{t_{k_{i}}}\neq\beta_{t_{k_{j}}}$ for $i\neq j$ without loss of generality.  Now, recall that $p_{\infty}$ is assumed to exist along the stochastic path of the test set. Thus, inspired by this assumption, let us further assume that $\lim\limits_{h\rightarrow
\infty}\frac{1}{h}%
%TCIMACRO{\tsum \limits_{j=0}^{h-1}}%
%BeginExpansion
{\textstyle\sum\limits_{j=0}^{h-1}}
%EndExpansion
P(A_{t_{k_{j}}+1}|$\ss $_{t_{k_{j}}})$ exists where $P(A_{t_{k_{j}}+1}|$\ss $_{t_{k_{j}}})=\beta_{t_{k_{j}}}$ or $P(A_{t_{k_{j}}+1}|$\ss $_{t_{k_{j}}})=\alpha$ along the path of the test set. Then, letting

\begin{center}
$\xi_{t_{k_{j}}}:=\begin{cases} 1 & P(A_{t_{k_{j}}+1}|$\ss $_{t_{k_{j}}}) = \alpha \\ 0 & P(A_{t_{k_{j}}+1}|$\ss $_{t_{k_{j}}}) = \beta_{t_{k_{j}}} \end{cases}$ \end{center}

\ \ \ \ (6) $\lim\limits_{h\rightarrow
\infty}\frac{1}{h}%
%TCIMACRO{\tsum \limits_{j=0}^{h-1}}%
%BeginExpansion
{\textstyle\sum\limits_{j=0}^{h-1}}
%EndExpansion
P(A_{t_{k_{j}}+1}|$\ss $_{t_{k_{j}}}) = \lim\limits_{h\rightarrow
\infty}\frac{1}{h}%
%TCIMACRO{\tsum \limits_{j=0}^{h-1}}%
%BeginExpansion
{\textstyle\sum\limits_{j=0}^{h-1}}
%EndExpansion
[$ $\xi_{t_{k_{j}}}\cdot P(A_{t_{k_{j}}+1}|$\ss $_{t_{k_{j}}}) + (1-\xi_{t_{k_{j}}})\cdot P(A_{t_{k_{j}}+1}|$\ss $_{t_{k_{j}}})$$]$ 

\ \ \ \ \ \ $= \alpha \cdot \lim\limits_{h\rightarrow
\infty}\frac{1}{h}%
%TCIMACRO{\tsum \limits_{j=0}^{h-1}}%
%BeginExpansion
{\textstyle\sum\limits_{j=0}^{h-1}}
%EndExpansion
$ $\xi_{t_{k_{j}}} + \lim\limits_{h\rightarrow
\infty}\frac{1}{h}%
%TCIMACRO{\tsum \limits_{j=0}^{h-1}}%
%BeginExpansion
{\textstyle\sum\limits_{j=0}^{h-1}} (1-\xi_{t_{k_{j}}})\cdot \beta_{t_{k_{j}}}$.

Thus, 

\ \ \ \ (7) $\lim\limits_{h\rightarrow
\infty}\frac{1}{h}%
%TCIMACRO{\tsum \limits_{j=0}^{h-1}}%
%BeginExpansion
{\textstyle\sum\limits_{j=0}^{h-1}}
%EndExpansion
P(A_{t_{k_{j}}+1}|$\ss $_{t_{k_{j}}}) = \alpha$, if and only if, $\lim\limits_{h\rightarrow
\infty}\frac{1}{h}%
%TCIMACRO{\tsum \limits_{j=0}^{h-1}}%
%BeginExpansion
{\textstyle\sum\limits_{j=0}^{h-1}} (1-\xi_{t_{k_{j}}})\cdot \beta_{t_{k_{j}}}=\alpha\cdot(1-\lim\limits_{h\rightarrow
\infty}\frac{1}{h}%
%TCIMACRO{\tsum \limits_{j=0}^{h-1}}%
%BeginExpansion
{\textstyle\sum\limits_{j=0}^{h-1}}
%EndExpansion
$ $\xi_{t_{k_{j}}})$. 

In other words, if Nature deviates from machine forecasts by $\beta_{t_{k_{j}}}$'s so that her deviating forecasts  on average satisfy (7) under (5), then $E$ $|p_{\infty}-$ $\alpha|$ $=0$ and thus the test set is truly guaranteed to be well-calibrated. But Nature then loses the repeated forecasting games along the path in the long run. So Nature has no reason to behave in this way with the true probability $P$- one. Let us then consider the following three cases:
\bigskip

\ \ \ \ (\textit{Case i}) $P ( P(A_{t+1} | ${\ss}$_t  ) = \alpha)=0$ at least $i.o.$

In this case, by \cref{lem:lemma2ndprob}, Nature observes machine forecasts $\alpha$ in each time $t_{k_{j}}$ whenever the machine predicts $P(A_{t+1} | ${\ss}$_t  )$ as $\alpha$. 

Now that $1=\limsup\limits_{t\rightarrow\infty}$ $P ( P(A_{t+1} | ${\ss}$_t  ) \neq \alpha)\leq P ( P(A_{t+1} | ${\ss}$_t  ) \neq \alpha$ at least $i.o. ),$

Nature would choose the deviating value $\beta_{t_{k_{j}}}$ in such a way that she would not allow (7) to hold with true probability $P$- one. Thus,

\ \ \ \ (8) $P $ $ (\lim\limits_{h\rightarrow
\infty}\frac{1}{h}%
%TCIMACRO{\tsum \limits_{j=0}^{h-1}}%
%BeginExpansion
{\textstyle\sum\limits_{j=0}^{h-1}} (1-\xi_{t_{k_{j}}})\cdot \beta_{t_{k_{j}}}=\alpha\cdot(1-\lim\limits_{h\rightarrow
\infty}\frac{1}{h}%
%TCIMACRO{\tsum \limits_{j=0}^{h-1}}%
%BeginExpansion
{\textstyle\sum\limits_{j=0}^{h-1}}
%EndExpansion
$ $\xi_{t_{k_{j}}}) $ $)\neq 1$. 

In other words, since Nature observes machine forecast $\alpha$ at every time, she would deviate each forecast $\alpha$ at $t_{k_{j}}$ in such a way that (8) holds in the end. Otherwise, $E$ $|p_{\infty}-$ $\alpha|$ $=0$, so Nature would lose in the long run. Therefore, we conclude due to (8) that $E$ $|p_{\infty}-$ $\alpha|$ $\neq 0$ in case (\textit{i}).
\bigskip
 
\ \ \ \ (\textit{Case ii}) $P ( P(A_{t+1} | ${\ss}$_t  ) = \alpha)=1$ at least $i.o.$

In this case, by \cref{lem:lemma2ndprob}, Nature moves first so the machine cannot fail to match $P(A_{t+1} | ${\ss}$_t)$. But then, 

\ \ \ \ $1=\limsup\limits_{t\rightarrow\infty}$ $P ( P(A_{t+1} | ${\ss}$_t  ) = \alpha)\leq P ( P(A_{t+1} | ${\ss}$_t  ) = \alpha$ at least $i.o. )=P ( P(A_{t+1} | ${\ss}$_t  ) \neq \alpha$ at most $f.o. )$, which contradicts (5). Therefore, we exclude case (\textit{ii}) under (5).
\bigskip

\ \ \ \ (\textit{Case iii}) $0<P ( P(A_{t+1} | ${\ss}$_t  ) = \alpha)<1$ at least $i.o.$

In this case, by \cref{lem:lemma2ndprob}, Nature moves simultaneously with the machine, so Nature has no reason to pick any particular $\beta_{t_{k_{j}}}\in\Re\lbrack0,1]$ at each $t_{k_{j}}$, for there exists no pure strategy Nash equilibrium. Hence any combination of $\{\beta_{t_{k_{j}}}\}_{j=0}^{\infty}$ is equally likely. Now, without loss of generality, let us fix $\alpha$ and $\xi_{t_{k_{j}}}$ for each $t_{k_{j}}$. Then we claim that 

\ \ \ \ (9) $P ($ $\frac{1}{h}%
%TCIMACRO{\tsum \limits_{j=0}^{h-1}}%
%BeginExpansion
{\textstyle\sum\limits_{j=0}^{h-1}} (1-\xi_{t_{k_{j}}})\cdot \beta_{t_{k_{j}}}\to c\alpha$ $)< P ($ $\frac{1}{h}%
%TCIMACRO{\tsum \limits_{j=0}^{h-1}}%
%BeginExpansion
{\textstyle\sum\limits_{j=0}^{h-1}} (1-\xi_{t_{k_{j}}})\cdot \beta_{t_{k_{j}}}\to c\alpha^{-} $ $)\leq1$ 

where $c=1-\lim\limits_{h\rightarrow
\infty}\frac{1}{h}%
%TCIMACRO{\tsum \limits_{j=0}^{h-1}}%
%BeginExpansion
{\textstyle\sum\limits_{j=0}^{h-1}}
%EndExpansion
$ $\xi_{t_{k_{j}}}$ for some fixed $c$, and $c\alpha\in C$ for some fixed $\alpha$, and some set $C$ such that $\forall x\in C$, $x\in\Re\lbrack0,1]$ but $C$ is countably infinite, and $c\alpha^{-}$ is any real number in the set $C$/$c\alpha$, the set $C$ without $c\alpha$.

First, recall that $\lim\limits_{h\rightarrow
\infty}\frac{1}{h}%
%TCIMACRO{\tsum \limits_{j=0}^{h-1}}%
%BeginExpansion
{\textstyle\sum\limits_{j=0}^{h-1}} (1-\xi_{t_{k_{j}}})\cdot \beta_{t_{k_{j}}}$ exists. Then, by definition, 

$\forall \epsilon>0$, $\exists $ $N_1<\infty$ such that $| \frac{1}{h}%
%TCIMACRO{\tsum \limits_{j=0}^{h-1}}%
%BeginExpansion
{\textstyle\sum\limits_{j=0}^{h-1}} (1-\xi_{t_{k_{j}}})\cdot \beta_{t_{k_{j}}}- c\alpha |< \epsilon, \forall h> N_1,$

$\forall \epsilon>0$, $\exists $ $N_i<\infty$ such that $| \frac{1}{h}%
%TCIMACRO{\tsum \limits_{j=0}^{h-1}}%
%BeginExpansion
{\textstyle\sum\limits_{j=0}^{h-1}} (1-\xi_{t_{k_{j}}})\cdot \beta_{t_{k_{j}}}- c_i\alpha^{-} |< \epsilon, \forall h> N_2. $ $ (1\neq i \in \mathbb{N})$

Now, letting $N=$ max$(N_1, N_i)$, $\forall \epsilon>0$, 

\ \ \ \ (10) $P(\{\omega\in\text{\ss }_{\infty}=%
%TCIMACRO{\tbigvee \limits_{j=0}^{\infty}}%
%BeginExpansion
{\textstyle\bigvee\limits_{j=0}^{\infty}}
%EndExpansion
\text{\ss }_{t_{k_{j}}}:$ $|$ $\frac{1}{h}%
%TCIMACRO{\tsum \limits_{j=0}^{h-1}}%
%BeginExpansion
{\textstyle\sum\limits_{j=0}^{h-1}}
%EndExpansion
$ $[ P(A_{t_{k_{j}}+1}|$\ss $_{t_{k_{j}}})=\beta_{t_{k_{j}}}]$ $- $ $c\alpha $ $|>\epsilon, \forall h>N \} ) < P($ $\bigcup\limits_{i=0}^{\infty} \{\omega\in$ \ss $_{\text{ }\infty}=%
%TCIMACRO{\tbigvee \limits_{j=0}^{\infty}}%
%BeginExpansion
{\textstyle\bigvee\limits_{j=0}^{\infty}}
%EndExpansion
$\ss $_{t_{k_{j}}}:$ $|$ $\frac{1}{h}%
%TCIMACRO{\tsum \limits_{j=0}^{h-1}}%
%BeginExpansion
{\textstyle\sum\limits_{j=0}^{h-1}}
%EndExpansion
$ $[ P(A_{t_{k_{j}}+1}|$\ss $_{t_{k_{j}}})=\beta_{t_{k_{j}}}]$ $- $ $c_i\alpha^{-} $ $|>\epsilon, \forall h >N\})\leq 1$. 

Therefore, we again obtain (8) by (10). Now, we consider all possible cases under (5), all of which lead to $E$ $| p_{\infty}-$ $\alpha) |$ $\neq0$. But this result is what we try to show in this proof anyway. Therefore, to continue to prove, let us accept that there exists such a set $E_A$ with true probability $P >0.$ 

Now, note that $E_{A}=\{\omega\in\text{\ss }_{\infty}=%
%TCIMACRO{\tbigvee \limits_{j=0}^{\infty}}%
%BeginExpansion
{\textstyle\bigvee\limits_{j=0}^{\infty}}
%EndExpansion
\text{\ss }_{t_{k_{j}}}:$ $ 1_{\{\omega\}}=1$ when $P(A_{t_{k_{j}}%
+1}|\text{\ss }_{t_{k_{j}}})=\beta\neq\alpha$ for all $t_{k_{j}}$'s along the
test set$\}\subset\{\omega\in$ \ss $_{\text{ }\infty}=%
%TCIMACRO{\tbigvee \limits_{j=0}^{\infty}}%
%BeginExpansion
{\textstyle\bigvee\limits_{j=0}^{\infty}}
%EndExpansion
$\ss $_{t_{k_{j}}}:$ $1_{\{\omega\}}=1$ when $|\lim\limits_{h\rightarrow
\infty}\frac{1}{h}%
%TCIMACRO{\tsum \limits_{j=0}^{h-1}}%
%BeginExpansion
{\textstyle\sum\limits_{j=0}^{h-1}}
%EndExpansion
P(A_{t_{k_{j}}+1}|$\ss $_{t_{k_{j}}})$ $-\alpha|\neq0$ for all $t_{k_{j}}%
$'s along the test set$\}$. Then, since $P(E_{A})>0$, $P($ $|\lim
\limits_{h\rightarrow\infty}\frac{1}{h}%
%TCIMACRO{\tsum \limits_{j=0}^{h-1}}%
%BeginExpansion
{\textstyle\sum\limits_{j=0}^{h-1}}
%EndExpansion
P(A_{t_{k_{j}}+1}|$\ss $_{t_{k_{j}}})$ $-\alpha|\neq0$ for all $t_{k_{j}}%
$'s along the test set $)>0$. Thus, since we found one subsequence of
$\{\frac{1}{h}%
%TCIMACRO{\tsum \limits_{j=0}^{h-1}}%
%BeginExpansion
{\textstyle\sum\limits_{j=0}^{h-1}}
%EndExpansion
P(A_{t_{k_{j}}+1}|$\ss $_{t_{k_{j}}})\}_{h=1}^{\infty}$ as such along the test set with true
probability $P>0$ and $p_{\infty}$ exists, $P(|\lim\limits_{k\rightarrow\infty}\frac{1}{k}%
%TCIMACRO{\tsum \limits_{j=0}^{k-1}}%
%BeginExpansion
{\textstyle\sum\limits_{t=0}^{k-1}}
%EndExpansion
P(A_{t+1}|$\ss $_{t})$ $-$ $ \alpha|\neq0$ along the test set$)>0$ for $\alpha\neq0$. Then, by the same reasoning as in 
\cref{lem:lemma2}, $E$ $|\lim\limits_{k\rightarrow\infty}\frac{1}{k}%
%TCIMACRO{\tsum \limits_{t=0}^{k-1}}%
%BeginExpansion
{\textstyle\sum\limits_{t=0}^{k-1}}
%EndExpansion
P(A_{t+1}|\text{\ss }_{t})-\alpha|$ $\neq0$. Now, by \cref{lem:lemma3}, we obtain
that $E$ $|p_{\infty}-\alpha|$ $\geq$ $E$ $|\lim\limits_{k\rightarrow\infty
}\frac{1}{k}%
%TCIMACRO{\tsum \limits_{j=0}^{k-1}}%
%BeginExpansion
{\textstyle\sum\limits_{t=0}^{k-1}}
%EndExpansion
P(A_{t+1}|\text{\ss }_{t})-\alpha|$ $\neq0$ when $p_{\infty}$ exists.
Clearly, when $p_{\infty}$ does not exist, $E$ $|p_{\infty}-$ $\alpha|$
$\neq0.$ 

Therefore, we conclude that if  $P (P(A_{t+1}|$\ss$_{t})\neq\alpha$ ${ }$ at least $i.o. ) >0$, then $E$ $|p_{\infty}-$ $\alpha|$ $\neq0.$ $Q.E.D.$

\textbf{Proof of \cref{thm:theorem5} } First, let us first note that with $P-$probability $>0,$ $P(A_{t+1}|$\ss $_{t})\neq1$ at least infinitely often for some event {$A_{t+1}$}. Otherwise, beyond the near future, all events $A_{t+1}$'s would certainly continue to occur, with $P-$probability one, and thus there would be no uncertainty about any $A_{t+1}$'s. Now, if this is the case, then we must stop here and simply conclude that no machine would be able to learn the true probability of any $A_{t+1},$ simply because there is no uncertainty for any machine to measure by the true probability in our world. Therefore, to continue to prove our main claim, we accept that $P(P(A_{t+1}|$\ss$_{t})\neq1$ at least $i.o. )>0$ for some event {$A_{t+1}$}. Now, let us consider the test set where $\alpha^{\ast}=1$. Then, along the stochastic path of this test set, $P(P(A_{t+1}|$\ss$_{t})\neq\alpha^{\ast}$ at least $i.o)>0$. Therefore, we found some $\alpha^{\ast}$ for which Nature is perverse with true probability $P>$ 0.

Now, suppose that, for any $\alpha$, $P (  P(A_{t+1}|\text{\ss }_{t})=\alpha\text{ })<1$ at least for infinitely many $t$'s. In other words, $P (  P(A_{t+1}|\text{\ss }_{t})\neq\alpha\text{ })>0$ at least $i.o.$ Then, $0 < \limsup\limits_{t\rightarrow\infty}$ $P ( P(A_{t+1} | ${\ss}$_t  ) \neq \alpha)\leq P ( P(A_{t+1} | ${\ss}$_t  ) \neq \alpha$ at least $i.o )$. Thus, by \cref{def:uniperverse}, Nature is uniformly perverse, which again means by \cref{def:perverse} that $P ($ Nature is perverse $) > 0$ for any $\alpha\in\Re[0,1].$ $Q.E.D.$

\textbf{Proof of \cref{thm:cannotlearn1}} Suppose that, for any $\alpha$, $P (  P(A_{t+1}|\text{\ss }_{t})=\alpha\text{ })<1$ at least for infinitely many $t$'s. Then, by \cref{thm:3cases} and \cref{thm:theorem5}, $E$ $|p_{\infty}-$ $\alpha| $
$\neq0$ and so $P($ $p_{k}\rightarrow$ $\alpha)\neq1$ for any $\alpha\in\Re[0,1]$ where $P$ is the true objective probability defined over \ss $_{\infty}=%
%TCIMACRO{\tbigvee \limits_{t=0}^{\infty}}%
%BeginExpansion
{\textstyle\bigvee\limits_{t=0}^{\infty}}
%EndExpansion
\text{\ss }_{t}$ and the expectation $E$ is taken with respect to this true
probability $P.$ Then, by \cref{thm:learncal}, the machine cannot learn the true objective
probability $P(A_{t+1}|$\ss $_{t}).$ $Q.E.D.$

\textbf{Proof of \cref{lem:tolerror}} Suppose that the machine effectively calculates $\Pi (A_{t+1} | ${\ss}$_t)$ as $\alpha$ with the goal of learning the true value of $P (A_{t+1} | ${\ss}$_t)$. Then, by the necessary condition for learning, the machine must return $\Pi (A_{t+1} | ${\ss}$_t)$ which is congruent to $P (A_{t+1} | ${\ss}$_t) = \alpha$, in order to achieve this goal. Now, suppose further that the machine calculates at the same time $\Pi ( \{P(A_{t+1} | $ {\ss}$_t ) \neq \alpha \}) \neq 0$. Then the machine tolerates error by \cref{def:tolerror}. 

However, by \cref{thm:learncal}, the machine cannot tolerate errors infinitely often to achieve this goal of learning for the following reason: for any $\alpha\in\Re\lbrack0,1]$, suppose that $\Pi (A_{t+1} | ${\ss}$_t) = \alpha$ but $\Pi ( \{P(A_{t+1} | ${\ss}$_t ) \neq \alpha\}) > 0$ infinitely often. Now, since it must be that $P (A_{t+1} | ${\ss}$_t) = \Pi (A_{t+1} | ${\ss}$_t) = \alpha$ to learn the true probability, it must also be by \cref{thm:learncal} that $P$ $(p_{k}\rightarrow$ $\alpha)$ = $\Pi$ $(p_{k}\rightarrow$ $\alpha)$ = 1. But now, by assumption, $\Pi ( \{P(A_{t+1} | ${\ss}$_t ) \neq \alpha\}) > 0$ infinitely often, which leads to that $0 < \limsup\limits_{t\rightarrow\infty}$ $\Pi ( \{P(A_{t+1} | ${\ss}$_t  ) \neq \alpha\})\leq\Pi ( \{P(A_{t+1} | ${\ss}$_t  ) \neq \alpha\}$ at least $i.o )$. But this contradicts $\Pi$ $(p_{k}\rightarrow$ $\alpha) = 1$ by the same reasoning as in the proof of \textbf{(Case 3)} in \cref{thm:3cases} while replacing $P$ by $\Pi$ and so the machine cannot learn the true probability by \cref{thm:learncal}. Therefore, the machine cannot tolerate errors infinitely often if the machine aims to learn the true probability. Since $\alpha$ was arbitrary in $\Re\lbrack0,1]$, let $\alpha = 0$, the desired result. $Q.E.D$

\textbf{Proof of \cref{lem:stoppingtime}} (\textit{i}) Proof of ``if'' part: suppose that there exists a stopping time $t_s < \infty$ for some forecast $\alpha_0$ such that $P(A_{\alpha_{0}}(t+1)|$\ss $_{t})=0,$ $ \forall t > t_s$, while there exists no stopping time for any other $\alpha\neq\alpha_0$ so that $P(A_{\alpha\neq\alpha_0}(t+1)|$\ss $_{t}) > 0$ at least infinitely often. Then, by the definition of $A_{\alpha_{0}}(t+1)$ and the law of iterated expectations, 

(11) \ \ \ \ $P(A_{\alpha_{0}}(t+1)) \searrow P ( \displaystyle{\lim_{t\to\infty} A_{\alpha_{0}}(t+1)})$, because $A_{\alpha_{0}}(t+1) \searrow\displaystyle{\lim_{t\to\infty} A_{\alpha_{0}}(t+1)}$.

Now that $\displaystyle{\lim_{t\to\infty} A_{\alpha_{0}}(t+1)}$ is the event that $P(A_{t+1}%
|$\ss $_{t})\neq\alpha_{0}$ at least $i.o.$ and so that the limit exists,

(12) \ \ \ \ $0= \displaystyle{\lim_{t\to\infty} P(A_{\alpha_{0}}(t+1))}=  P(P(A_{t+1}%
|$\ss $_{t})\neq\alpha_{0}$ at least $i.o.)$ for $\alpha_0$.

Also, in the same logic as for $\alpha_0$,

(13) \ \ \ \ $0 < \displaystyle{\lim_{t\to\infty} P(A_{\alpha} (t+1))}=  P(P(A_{t+1}%
|$\ss $_{t})\neq\alpha$ at least $i.o.)$ for any $\alpha\neq\alpha_0.$ 

Thus, by \cref{def:selperverse}, Nature is selectively perverse. 

(\textit{ii}) Proof of ``only if'' part: suppose that Nature is selectively perverse. Then, by \cref{def:selperverse}, there must exist some $\alpha_0$ such that $P(P(A_{t+1}%
|$\ss $_{t})\neq\alpha_{0}$ at least $i.o.)=0$. Now, for \textit{reductio}, suppose that for any such $\alpha_0$ there exists no stopping time $t_s$ so that $P(A_{\alpha_{0}}(t+1)|$\ss $_{t})>0 $ at least infinitely often. In other words, Nature keeps changing her mind infinitely often between perversity and non-perversity or Nature keeps being perverse all the way long. Then, by law of iterated expectation, $P(A_{\alpha_{0}}(t+1)) > 0 $ at least infinitely often, which contradicts the selective perversity of Nature by the same reasoning as in (13). $Q.E.D.$

\textbf{Proof of \cref{lem:learnselfassured}} For any given $\alpha_0$ with which Nature is not perverse with true probability $P$-one, there exists $t_s < \infty$ for this $\alpha_0$ by \cref{lem:stoppingtime}. Now, by assumption, machines learn that $P(A_{\alpha_0}(t+1)|$\ss $_{t})=0$ $\forall t > t_s$. Thus,  $\Pi(A_{\alpha_0}(t+1)|$\ss $_{t})=0$ $\forall t > t_s$ by the necessary condition for learning. Then, by \cref{lem:tolerror} and the same reasoning as (11) in the proof of \cref{lem:stoppingtime}, $\Pi ( P(A_{\alpha_0}(t+1)|$\ss $_{t})=0, \forall t > t_s) = 1$. $Q.E.D.$ 

\textbf{Proof of \cref{cor:corollary1}} (\textit{i}) Suppose that Nature is selectively perverse so that $P(A_{\alpha_0}(t+1)|$\ss $_{t})=0$ $\forall t > t_s$ for some $\alpha_0$ by \cref{lem:stoppingtime}. However, since the machine is assumed not to be self-assured that the stopping time $t_s$ arrives for that $\alpha_0$, the machine cannot learn that $P(A_{\alpha_0}(t+1)|$\ss $_{t})=0$ $\forall t > t_s$ by \cref{lem:learnselfassured}. 

(\textit{ii}) Now, note that if the machine learns $P(A_{t+1}|$\ss $_{t})$ as $\alpha
_{0}$, the machine also learns that $P(P(A_{t+1}|$\ss $_{t})\neq\alpha_{0}$ at least $i.o. )=0$ in the following way: first, by \cref{thm:learncal} and (Case 3) in \cref{thm:3cases}, machine learning of the true probability $P(A_{t+1}|$\ss $_{t})$ as $\alpha
_{0}$ mathematically implies that $P(P(A_{t+1}|$\ss $_{t})\neq\alpha_{0}$ at least $i.o. )=0$. Thus, once the machine learns the true probability $P(A_{t+1}|$\ss $_{t})$ as $\alpha_{0}$, it \textit{cannot fail to effectively calculate} the
true probability $P(A_{\alpha_0}(t+1))$ as $0$, following \cref{thm:learncal} and (Case 3) in \cref{thm:3cases} as instructions. Then, by \cref{def:learn}, the machine \textit{learns} that $P(A_{\alpha_0}(t+1))=0$ in particular $\forall t>t_s$, so that $P(A_{\alpha_0}(t+1)|$\ss$_t)=0$ $\forall t>t_s$ while following law of iterated expectation as instruction. However, as we proved it in (\textit{i}), the machine cannot learn that $P(A_{\alpha_0}(t+1)|$\ss $_{t})=0$ $\forall t > t_s$. Hence we conclude that the machine cannot learn the true objective probability $P(A_{t+1}|$\ss $_{t})$ as $\alpha
_{0}$ either. $Q.E.D$.

\textbf{Proof of \cref{lem:condcor1}} Suppose that the machine is not self-assured of the stopping time $t_s$ for $\alpha_0$. Then, 

(14) \ \ \ \ $\Pi ( P(A_{\alpha_0}(t+1)|$\ss $_{t})=0, \forall t > t_s) \neq 1$.

Now that $\displaystyle{\lim_{t\to\infty} P(A_{\alpha_{0}}(t+1))}=  P(P(A_{t+1}%
|$\ss $_{t})\neq\alpha_{0}$ at least $i.o.)$ for this $\alpha_0$,

(15) \ \ \ \ $ \Pi ( $ $ P( $ $ P(A_{t+1} | ${\ss}$_t  )\neq \alpha_0$   at least $i.o. )= 0 ) \neq1$.

Then, since $\limsup\limits_{t\rightarrow\infty}$ $P( $ $ P(A_{t+1} | ${\ss}$_t  ))\neq \alpha_0 )\leq P( $ $ P(A_{t+1} | ${\ss}$_t  )\neq \alpha_0$   at least $i.o. ) = 0$,

(16) \ \ \ \ $ \Pi ( $ $ P( $ $ P(A_{t+1} | ${\ss}$_t  ) =  \alpha_0) = 1$ $ \forall t > t^{\ast} ) \neq 1$, for some $t^{\ast}<\infty$. 

Now, note that along the stochastic path considered in \cref{cor:corollary1}, $P(P(A_{t+1}%
|$\ss $_{t})\neq\alpha_{0}$ at least $i.o.)=0$ $\forall t > t_s$. Now, for this $\alpha_0$,  

(17) \ \ \ \ $\limsup\limits_{t\rightarrow\infty}$ $P ( P(A_{t+1} | ${\ss}$_t  ) \neq \alpha_0)\leq P ( P(A_{t+1} | ${\ss}$_t  ) \neq \alpha_0$ at least $i.o )=0$ 

Therefore, without loss of generality, letting $t^{\ast} \geq t_s$ with $t^{\ast} < \infty$,

(18) \ \ \ \ $P ( P(A_{t+1} | ${\ss}$_t  ) = \alpha_0) = 1,$ $\forall t > t^{\ast} \geq t_s$ with $t^{\ast} < \infty$. 

Then, without loss of generality, let $P ( P(A_{t+1} | ${\ss}$_t  ) = \alpha_0) = 1$ at $t^{\ast}$+1 by (18). Thus, (16) and (18) lead to the desired result by \cref{lem:lemma2ndprob}. $Q.E.D$ 

\textbf{Proof of \cref{thm:learnselfassured}} Suppose that the machine learns the true probability. Since the machine cannot learn if Nature is uniformly perverse, Nature must then be selectively perverse so that the stopping time $t_s$ exists by \cref{lem:stoppingtime}. Then, by the (\textit{ii}) part of \cref{cor:corollary1} and \cref{lem:learnselfassured}, the machine is self-assured of the stopping time $t_s$ when $t_s$ exists. We now finish the proof of \cref{thm:learnselfassured} by showing that if the machine learns the true probability, the machine is not self-assured of the stopping time $t_s$ when such $t_s$ does not exist.

Suppose that the machine is self-assured of the stopping time $t_s$ even though such $t_s$ does not exist. The machine is then wrong about $t_s$, so it cannot learn the true probability along the path where $P (  A_{\alpha} (t+1) | ${\ss }$_{t} )  >0 $ at least $i.o.$ for the following reason: first, by \cref{lem:stoppingtime}, with true probability $P>0$, Nature is perverse to the forecast $\alpha$ along the path where there is no stopping time $t_s$. Thus, $P(P(A_{t+1}%
|$\ss $_{t})\neq\alpha$ at least $i.o.)>0$ for such forecast $\alpha.$ Then, by the (Case 3) of \cref{thm:3cases} and then \cref{thm:learncal}, the machine cannot learn that $\alpha$. In other words, the world does not exist in the way that Nature allows the machine to learn the true probability. Notwithstanding, the machine has a wrong belief about the stochastic path of the true probability, and so cannot learn the true probability. $Q.E.D.$  

\textbf{Proof of \cref{thm:selfassureddo}} Suppose that the machine is self-assured of stopping time $t_s$ along the path where, for any given $\alpha_0$, $P(A_{\alpha_0}(t+1)|${\ss}$_{t})=0$ $\forall t>t_s$. Then, along this path, the machine obtains

$\Pi (P(A_{\alpha_0}(t+1)|${\ss}$_{t})=0$ $\forall t>t_s)=1$ and so $\Pi (A_{\alpha_0}(t+1)|${\ss}$_{t})=0$ $\forall t>t_s$ by \cref{lem:tolerror}.

Now, by the definition of $A_{\alpha_0}(t+1)$ and \cref{lem:tolerror} again, 

$\Pi (A_{t+1}|${\ss}$_{t} )= \alpha_0$,  $\forall t>t^{\ast}>t_s$ for some $t^{\ast}<\infty$

Note also that $P (A_{t+1}|${\ss}$_{t} )= \alpha_0$,  $\forall t>t^{\ast}>t_s$ for some $t^{\ast}<\infty$ along this path.
\bigskip

(19) \ \ \ \ $P(A_{t+1}| ${\ss}$_{t}) = \alpha_0 = \Pi(A_{t+1}| ${\ss}$_{t})$,  $\forall t > t^{\ast}$ with $t^{\ast}<\infty$.
\bigskip

Then, as in \cref{thm:learncal}, we can construct a test set along the stochastic path by the assessed $\alpha_0$ as a selection criterion by (19). This test set is also truly guaranteed to be well-calibrated.

Thus, from this test set along the path, the machine obtains the following by \cref{lem:lemma2} and \cref{lem:lemma3},

\bigskip
(20) \ \  $P$ $(\lim\limits_{n\rightarrow\infty}\frac{1}{n}\sum\limits_{t=t^{\ast}}^{t^{\ast} + n}P(A_{t+1}|${\ss} $
_{t})= \alpha_0) = 1$ if and only if $P$ $(\lim\limits_{n\rightarrow\infty}\frac{1}{n}\sum\limits_{t=t^{\ast}}^{t^{\ast} + n} 1_{\{A_{t+1}\}}$ $=$ $\alpha_0)=1$ 
\bigskip

Now, let us gather the sequence of $\{A_{t+1}\}_{t=t^{\ast}}^{\infty}$ along the path and call this set a population. The machine then effectively calculates the true probability $P(A_{t+1} | ${\ss}$_t)$ as $\alpha_0$ by the empirical distribution out of this population by (20), which satisfies $(i)$ in \cref{def:direcobs}. Also, this effective calculation of the empirical distribution must be successful in returning the true probability $P(A_{t+1} | ${\ss}$_t)$, for $\frac{1}{n}\sum\limits_{t=t^{\ast}}^{t^{\ast} + n}P(A_{t+1}|${\ss} $
_{t})$ in the right-hand side of (20) is equal to $P(A_{t+1}|${\ss}$_{t}), \forall n$ and $\forall t>t^{\ast}$ by (19), which satisfies $(ii)$ in \cref{def:direcobs}. Therefore, by \cref{def:direcobs}, the machine directly observes the true probability $P(A_{t+1} | ${\ss}$_t)$ as $\alpha_0$. $Q.E.D$

\textbf{Proof of \cref{thm:dolearning}} (\textit{i}) Proof of ``if'' part: follows directly from \cref{thm:learnselfassured} and \cref{thm:selfassureddo}.

(\textit{ii}) Proof of ``only if'' part: suppose that the machine directly observes the true probability $P$$(  A_{t+1}|${\ss }$_{t})$ as $\alpha$ from the given population $S$ at some time $t^{\ast}$. The machine then effectively calculates $\Pi$$(  A_{t+1}|${\ss }$_{t})$ as $\alpha$ at $t^{\ast}$, while adopting the following as an instruction: recall that the given set $S$ consists of the sequence of events $A_{t+1}$'s, $\{A_{t+1}\}_{t=0}^{k-1}$ with $k$ potentially infinite. Since the set $S$ is available in principle to the machine by the part (\textit{i}) of \cref{def:direcobs}, there must exist some rule on how to collect the available set of events $\{A_{t+1}\}_{t=0}^{k}$. Then let the machine build up the population $S$ by collecting events while following the rule on how-to. Now, once collected by the machine to constitute the set $S$, it must have been observed whether each event has a certain attribute of interest or not, and so a value of the indicator variable $1_{\{A_{t+1}\}}$ must have been assigned accordingly to each event $A_{t+1}$ by the machine. Then, let the machine calculate $\Pi$$(  A_{t+1}|${\ss }$_{t})$ as $\alpha$ $= \frac{1}{k}\sum\limits_{t=0}^{k-1} 1_{\{A_{t+1}\}}$. Therefore, the machine \textit{effectively calculates} $\Pi$$(  A_{t+1}|${\ss }$_{t})$ as $\alpha$.

Furthermore, note that $ \frac{1}{k}\sum\limits_{t=0}^{k-1} 1_{\{A_{t+1}\}}$ is defined to be $P$$(  A_{t+1}|${\ss }$_{t})$ at $t^{\ast}$ by the part (\textit{ii}) in \cref{def:direcobs}. The machine then \textit{cannot fail} to compute $P$$(  A_{t+1}|${\ss }$_{t})$ as $\alpha$ from the population $S$. Therefore, the machine learns the true probability $P$$(  A_{t+1}|${\ss }$_{t})$ as $\alpha$ by \cref{def:learn}. $Q.E.D.$

\textbf{Proof of \cref{cor:successcriterion}} Let us first define what we mean by ``most of the time'' in the success criterion (1). by \cref{lem:lemma2} and \cref{thm:3cases}, machines cannot satisfy the calibration property when the test set is constructed by the selection criterion of an assessed probability $\alpha$ if $P(P(A_{t+1}|$\ss$_{t})\neq\alpha$ at least $i.o.) >0$. Therefore, in order to learn, the machines must return the correct calculations except a finite number of times out of infinite opportunities to learn. Thus, ``most of the time'' in the Success Criterion (1) should be ``all but finitely often out of infinite opportunities to learn,'' which means that machines must be correct not just infinitely often while being wrong that often. 

Now suppose that the machine is correct most of the time when the machine aims to learn the true probability $P(A_{t+1}|$\ss$_{t})$. Then, by the \textbf{(Case 1)} in \cref{thm:3cases},  $P(P(A_{t+1}|$\ss$_{t})\neq\alpha$ at most $f.o.)=1$. Thus, there exists a stopping time $t_s$ because $P(P(A_{t+1}|$\ss$_{t})\neq\alpha$ at least $i.o.)=0$ if and only if there exists a stopping time $t_s$ for any machine forecast $\alpha$ by \cref{lem:stoppingtime}. Furthermore, suppose that the machine is self-assured that it is correct most of the time. Then, again by \cref{lem:stoppingtime}, $\Pi($ there exists a stopping time $t_s) =1.$ Thus, if the machine satisfies the Success Criterion (1), then it satisfies the condition of \cref{thm:selfassureddo}. Therefore, if the machine satisfies the Success Criterion (1), it can learn the true probability by \cref{thm:selfassureddo} and \cref{thm:dolearning}. $Q.E.D.$

\section{Some Literature for the Necessary Condition in Sec. 3.2}

There has been a large literature in logic and economics whose discussion implies when a machine holds a true belief in the probabilistic proposition $A_{p}$. For example, while defining the concept of rationality in the economics model, \cite{SargentC:08, SargentC:09}, \cite{Sandroni:00}, \cite{BlumeE:06, BlumeE:08} and many others stipulate that an agent is \textit{rational} when his/her partial beliefs are \textit{correct} in the sense that his/her subjective probability distributions are congruent to the true probability distribution which Nature identifies as such. In other words, this means that a machine holds such a true belief in $A_{p}$ when it is rational, which entails that its subjective probability $\Pi$ is equal to the true objective probability $P.$

Also, in probabilistic logic, \cite{Nilsson:86}, \cite{HalpernF:94}, and many others follow the probabilistic version of the Tarskian semantic theory of truth in the following way: a formula describing the subjective probability of an agent is \textit{true} when the agent's probability assignment corresponds to what the sentence in fact represents. For example, in \cite{HalpernF:94}, a formula like $w_{i}(\varphi)\geq2w_{i}(\psi)$ is true if, according to the probability assignment of the agent $i$, the event $\varphi$ is at least twice as probable as $\psi$. Now, if we extend this idea to the true objective probability $P$ if any, a formula such as $w_{i}(\varphi)=w(\varphi),$ where $w_{i}$ denotes the probability operator of the agent $i $ and $w$ does that of Nature, is true when, according to the assignment of the agent $i$'s probability, the event $\varphi$ is as probable as what Nature assigns on $\varphi$ as the true probability value in our world.

It deserves to note from the economics literature when it becomes true that agent $i$'s partial belief on the event $\varphi$ has a degree $w_{i}%
(\varphi)$ which corresponds to the true objective probability $w(\varphi)$. This is indeed true when the subjective probability of the agent $i$, $w_{i}(\varphi)$ is in congruence with the true objective probability $w(\varphi)$, which again makes the formula $w_{i}(\varphi)=w(\varphi)$ true. Therefore, the condition for any agent to be rational (or rational machine in our context) in economics is equivalent to the truth condition for the formula in probabilistic logic. 

\section{Justifications for the Three Assumptions}

\textbf{\cref{ass:assumption1}} \ss $_{t}$'s in $P(A_{t+1}|$\ss $_{t})$ are the set of all the \textit{true facts} up to time $t$.

In other words, \ss $_{t}$ is the historical path of true facts up to time $t$. To recognize that \cref{ass:assumption1} is reasonable, recall that we are handling with objective probability true to our world. Therefore, its condition must also be true in our world. Otherwise, $P(A_{t+1}|$\ss $_{t})$ cannot represent the true probability according to which the actual data are realized in our world. For example, if there works some special gravity force on Mars and so a fair coin lands on its edge as equally likely as on its head or tail, then the probability of the coin landing on the head \textit{conditional on} this hypothesis will be $\frac{1}{3}$. However, if such a special gravity force \textit{actually} does not exist on Mars, this conditional probability $\frac{1}{3}$ cannot be true either, because its data would not be realized according to the probability of $\frac{1}{3}$ in our world.

\textbf{\cref{ass:assumption2}} No further knowledge requirement is imposed on the condition \ss $_{t}$.

To recognize that \cref{ass:assumption2} is reasonable, note the following: If
\ss $_{t}$ is the set of \textit{known} facts, then $P(A_{t+1}|$\ss $_{t})$ can vary from person to person, as the set of events known to each person may be different, depending on who possesses what information. In order for $P(A_{t+1}|$\ss $_{t})$ to be objective, however, $P(A_{t+1}|$\ss $_{t})$ should not depend on each person. Therefore, we require that \ss $_{t}$ consist of true facts, not necessarily knowledge.

\textbf{\cref{ass:assumption3}} Once a probability of an event type $E$ is established, its associated event tokens $E_{t_{k}}$'s \textit{occur}\ at some infinite subsequence of time $t_{k}$' s, so that $P(E_{t_{k}})$ does not vanish to zero as $t_{k}\rightarrow\infty$.

Here, ``event token'' refers to the event that ever occurs at some specific time and place, while ``event type'' refers to the abstract object with no specific space-time location. For example, cloudy weather in Denver is an abstract event type $E$ with no time subscript, while cloudy weather in Denver on 29 May 2024 is a particular event token $E_{t_{0}}$. Some literature (\textit{e.g.} \cite{Halpern:16}) deals mainly with probability of token events, while some literature (\textit{e.g.} \cite{Maher:10}) deals mainly with probability of type events. \cref{ass:assumption3} establishes a connection between the probabilities of these two kinds of events.

In order to recognize that \cref{ass:assumption3} is reasonable, consider now the following example: suppose that we try to predict the probability that some person $i$ suffers from lung cancer caused by his/her smoking habit. As we discussed in the Introduction, this causal probability is objective, which is relevant to our discussion. Then, as long as the probability of the event type of having lung cancer from smoking is allowed to be considered for forecasting, we require that the true probability of the associated event tokens for some persons $i$'s should not be completely zero from some time $t_{0}<\infty$ onward. In other words, although the true probability of such event tokens is allowed to be intermittently zero, the
probability of the associated event tokens should not vanish to zero as $k\rightarrow\infty$.

It might be pointed out that a particular person, say Mary, will die some time in the future, and that it will not make sense to consider the probability of Mary's suffering from lung cancer after that time any more. However, unless all generations of our human beings suddenly become extinct in the near future, we can consider the true probability of this event token at least for some person $i$ at each time $t$. Hence it would make sense to forecast the probability of such an event token in each specific case, as $t\rightarrow\infty.$

\section{More Detailed Remarks}

\textbf{\cref{rem:truepro}} Now, let $ \mathcal{F} $ be the sigma-field generated by $\Omega$ and $\omega^{t}=(S_{0}^{-1}(s_{0}),$ $\ldots,S_{t}^{-1}(s_{t}),$ $\Omega_{t+1},\Omega_{t+2},$ $\ldots)\in\Omega$ denote a partial history through date $t$. Then, for any probability measure $p_{t}$ on $ \mathcal{F}_{t},$ $p_{t}(\omega^{t})$ becomes the (marginal) probability of the partial history, and each $\omega^{t}$ is assumed to be $\mathcal{F}_{t}$-measurable. Note then that $p_{t}{(\omega^{t})=\textstyle\prod\limits_{\tau = 1}^{t}} p(\omega_{\tau}| \mathcal{F}_{\tau-1})$ for any $t,$ and so $p_{t}(\omega^{t})=p(\omega_{t}| \mathcal{F}_{t-1})p_{t-1}(\omega^{t-1}).$ Furthermore, when $s_t$ is only either 0 or 1, $S_t(\omega_t)$ becomes an indicator function for an event $\{\omega_t\}.$ Then, provided that there indeed exists any true objective probability $P$, $p(\{\omega_t\}|%
%TCIMACRO{\tciFourier}%
%BeginExpansion,
\mathcal{F}%
%EndExpansion
_{t-1})=P(\{\omega_t\}|%
%TCIMACRO{\tciFourier}%
%BeginExpansion
\mathcal{F}%
%EndExpansion
_{t-1})$ $=E(S_t(\omega_t) = 1|%
%TCIMACRO{\tciFourier}%
%BeginExpansion
\mathcal{F}%
%EndExpansion
_{t-1})$ where the expectation $E$ is taken with respect to this true probability $P$.

For example, let $S_{t}$ be an $\mathbf{i.i.d}$\textbf{.} random variable whose value is $1$ if the event $\{\omega_{t}\}$ occurs at $t$ and $0$ otherwise. Then, $X_{n}=\sum\limits_{k=1}^{n}S_{k}$ will be the number of events that have occurred up to time $n$. Since $S_{t}$ is $\mathbf{i.i.d}%
$\textbf{.}, $p(\{\omega_{t}\}|%
%TCIMACRO{\tciFourier}%
%BeginExpansion
\mathcal{F}%
%EndExpansion
_{t-1})$ is same as $P(\{\omega_{t}\})$ across time. Now, let $\lim
\limits_{n\rightarrow\infty}\frac{X_{n}}{n}=$ $\lim\limits_{n\rightarrow
\infty}\frac{1}{n}\sum\limits_{k=1}^{n}S_{k}$ be the ratio of events that ever occur. Then, provided that this limit indeed exists, the dominated convergence theorem and Fubini's theorem imply that $E\{\lim\limits_{n\rightarrow\infty }\frac{1}{n}\sum\limits_{k=1}^{n}S_{k}\}=P(\{\omega_{t}\})$. Thus, in the $\mathbf{i.i.d}$\textbf{.} case, we can derive that with the true probability $P-$ one, the true objective
probability of the event $\{\omega_{t}\}$ is\textit{ } the limiting relative frequency which is objective.

By stipulating that the true objective probability follows the rule on how Nature generates each actual data point, we emphasize that the true probability here is something objective, not subjective, but no more or no less than that. ``Nature'' is just a metaphor for describing the relationship of
true probability with our objective world. Adopting the widely accepted statistical notion of a data-generating process, we intend to use the term ``Nature'' to refer to whatever is supposed to govern the underlying true objective process to generate the actual data. Given that Nature is simply a metaphor, it is important to emphasize that, in order to prove the possibility or the impossibility of machine learning on the true objective probabilities, we do not need to commit ourselves to whether there really \textit{exists} such a thing as a true objective process: probability might be merely something \textit{subjective} which has nothing to do with ``Nature.'' If that is the case, then we conclude that no machines can learn the true objective probabilities simply because there exist no such things as true probabilities for machines to learn.

\textbf{\cref{rem:probinterpretation}} The standard theory of subjective probability was first developed by Ramsey and then further by De Finetti and Savage. Subjective probability is designed to represent a \textit{degree of belief} possessed by a subject, say
some person. Here, two words, \textit{degree }and\textit{\ belief}, deserve to be noted. First, subjective probability represents some aspects of \textit{belief}. However, belief is an inner thought that, in principle, resists a direct observation, while probability quantification requires
measurability. Note that the easiest method of measurement is by observation. Thus, in order for the degree of belief to be quantified as a probability measure, it works well if the unobservable is made observable. Here comes in the relationship between unobservable belief and observable action: belief
causes action. According to \cite{Ramsey:31}, the strength of our \textit{belief} can be judged in relation to how we should \textit{act} in hypothetical situations. Given a preferential system on the lotteries of a set of conditions, the choice \textit{ action} under hypothetical circumstances will \textit{ reveal} the degree of \textit{belief} of some relevant agent. In this vein, subjective probability represents whatever is in any one's mind upon anything as long as his/her belief system is coherent, and thus can be even assigned to what is merely imagined. For instance, while arguing for \textit{cogito, ergo sum,} \cite{Descartes:08} imagined some evil spirit that has devoted all its efforts to deceiving him. Then, Descartes can assign some value of subjective probability to such imagination on the evil spirit, according to how likely it is to him that such imagination can be realized in this world, as long as Descartes' belief system is coherent.

Second, it is assumed that the \textit{degree} of belief ranges between 0 and 1. For example, your belief that there will be rain tomorrow has a degree strictly less than $1$ and thus is called a partial belief, because you have some unconfidence on future events. In addition to this quantitative usage of the term ``belief'', however, there is another categorical usage: ``belief'' refers to the proposition that something is the case or that something is not the case, or none of them. For instance, your belief in the Moorean
fact that here is one hand represents either the case or not, or it is on suspension. Compared to partial belief, this qualitative belief is called belief \textit{simpliciter}. As the term ``belief'' has these two faces, gradational quantitative and categorical qualitative ones, \textit{numerical degrees} are assigned to partial belief, while \textit{truth values} are assigned to belief \textit{simpliciter.} In this paper, we abbreviate belief \textit{simpliciter} by ``belief'' and denote partial belief by ``partial belief'' as it is.

In contrast, objective probability, if any, is what must be determined by objective features of the world that do not vary from person to person. Following \cite{Nagel:39} and \cite{Carnap:63}, we list \textit{chance}, \textit{logical probability}, and \textit{relative frequency} as exhaustive examples of objective probability. The best way to clarify these concepts is
to consider their examples. Following \cite{Maher:10}, for example, suppose that a coin has the same face on both sides, that is, two-headed or two-tailed. Provided further that it is completely uncertain what face value, head or tail, the coin has on both sides, the \textit{chance} of getting head when
tossed is 1 or 0, while its \textit{logical probability} is $\frac{1}{2}$. Furthermore, when the coin is tossed infinitely often, its \textit{relative frequency} surely converges to 1 or 0.

Here, the chance is either 1 or 0, depending on what our world is like, namely, whether the coin is indeed two-headed or two-tailed. Therefore, the chance is objective in the sense that it depends on real features of the coin, not on any personal inner thought. On the other hand, the logical probability is $\frac{1}{2},$ because it is logically implied from the given
conditions that the coin has the same face value on both sides, but that whether it is two-headed or two-tailed is completely uncertain. Therefore, logical probability is also objective in the sense that it depends on the logical features of our world, not on us. Clearly, the relative frequency is
what our world turns out to be, not whatever we believe. However, no matter what interpretation of probability is adopted among these three kinds, it is important to note that the true objective probability $P$ in \cref{def:truepro} is a mathematical object that is supposed to represent any of them as long as they satisfy the Kolmogorov axioms.

\textbf{\cref{rem:learncal}} It should be noted that \cref{thm:learncal} is our building block to prove when a machine cannot learn the true probability, because $p_{\infty}$ in \cref{thm:learncal} denotes the limiting relative frequency along the test set, the representative true objective probability. We do not consider any limiting behavior of the relative frequency outside the test set, because learning as $\alpha$ per se is not possible outside the test set by the necessary condition for learning in Section 3.2. Therefore, if it is shown to be impossible that with $P-$probability one$,$ $p_{k}\rightarrow$ $\alpha$ along the stochastic path of the test set collected by the assessed $\alpha$, then it is derived from \cref{thm:learncal} that the machine cannot learn the true probability. 

Now, note that $P ($$p_{k}\rightarrow $ $\alpha$ ) $ = 1$ if and only if for any $\epsilon >0,$ $\lim\limits_{n\rightarrow \infty
}P(\sup\limits_{m\geq n}$ $|$ $p_{m}-\alpha$ $|$ $<\epsilon ) = 1.$ Thus, if the machine learns, then for all $\epsilon >0$ that are small enough, $\lim\limits_{n\rightarrow \infty
}P(|$ $p_{n}-\alpha $ $|$ $<\epsilon ,$ $|p_{n+1}-\alpha |<\epsilon ,\ldots) = 1$, which is $\lim\limits_{n\rightarrow \infty }P(p_{n}=\alpha ,$ $p_{n+1}=\alpha
,\ldots ) = 1.$ Thus, \cref{thm:learncal} is not committed to what the machine engages in by the first $n-1$ number of data while ``learning''. This concept of machine learning is flexible enough to allow for some finitely few potential errors where $p_{t}\neq\alpha$ $\forall t < n$ so that $P(A_{t+1}|$\ss$_{t})\neq\alpha$ $\forall t < n$ while processing the data to learn.      

\textbf{\cref{rem:count1}}
Indeed, it may well be argued against the \cite{Oakes:85} Counterexample that, although it could be imagined so, Nature \textit{actually} never behaves in that way. There is no reason why Nature is so perverse that she generates data in such a deviating way. The true objective probability of Nature being perverse may be simply zero. Then, \cref{thm:dawid} and \cref{thm:impossdawid} do not necessarily imply that a machine cannot learn the true probability.

\cref{thm:impossdawid} shows only to the extent that if a machine can imagine such a counterexample, and thus it sincerely believes in such possibility, then its subjective probability of long-run mis-calibration is not zero. But recall the Descartes' Demon case from Section 3.1. A simple possibility of imagination
does not necessarily imply a \textit{real} possibility, namely that the true objective probability of it occurring in the actual world is not zero. \cref{thm:dawid} and \cref{thm:impossdawid} show only that if a machine cannot exclude such a counterexample, it cannot be self-assured to be well-calibrated with its own subjective probability $1$. However, recall that there exists an asymmetric relation between subjective and objective probabilities: objective probability binds subjective probability, but not necessarily vice versa. Thus, if the \textit{true} probability of Nature's perversity is \textit{proven} to be zero, the machine can exclude such a possibility, and so its subjective probability on Oakes' counterexample will be zero as well. Then, from this it is derived neither that the machine cannot be self-assured to be well-calibrated nor that it cannot be truly guaranteed to be so, which implies that the impossibility of machine learning does not necessarily follow from \cref{thm:learncal}.

Later by \cref{thm:theorem5}, we prove that such an imagined possibility of Nature's being perverse is a real one if the true probability is not observable. Meanwhile, we will also prove mathematically how \cite{Oakes:85} Counterexample paralyzes Dawid's \cref{thm:dawid}, which amounts to the proof of \cref{thm:impossdawid}. Note that if the true probability indeed escapes from the machine's forecast just as in \cite{Oakes:85}, \cref{thm:dawid} breaks down: \cref{thm:dawid} critically relies on the martingale property of $%
%TCIMACRO{\tsum \limits_{t=1}^{k}}%
%BeginExpansion
{\textstyle\sum\limits_{t=1}^{k}}
%EndExpansion
X_{t}$ given \ss $_{k-1}$ where $X_{t}=(%
%TCIMACRO{\tsum \limits_{j=1}^{t}}%
%BeginExpansion
{\textstyle\sum\limits_{j=1}^{t}}
%EndExpansion
\xi_{j})^{-1}\cdot\xi_{t}(Y_{t}-\hat{Y}_{t})$, which was from $E(X_{k}%
|$\ss $_{k-1})=0.$ This martingale property, however, breaks down when
$P(A_{t+1}|$\ss $_{t})=E(Y_{t+1}|$\ss $_{t})\neq\hat{Y}_{t+1}=\Pi(A_{t+1}%
|$\ss $_{t})$ for all $t$. Note that \cite{Dawid:82} takes it for granted that
$E(Y_{t+1}|$\ss $_{t})=\Pi(A_{t+1}|$\ss $_{t})=\hat{Y}_{t+1}$ for all $t$. Therefore, if we relax this assumption, we can prove mathematically how \cite{Oakes:85} works against \cite{Dawid:82}, which will be shown from $($\textbf{Case 2}$)$ in the proof of \cref{thm:3cases}.

\textbf{\cref{rem:lemma23}} Regarding \cref{lem:lemma2} and \cref{lem:lemma3}, it deserves to note the following two things: first, note that we do not require any standard assumption such as the stochastic process to be $\mathbf{i.i.d.}$ along the historic path of the test
set and so that $P(A_{t+1}|$\ss $_{t})$ can vary along the path. Note also that unlike \cite{BlumeE:06, BlumeE:08}, etc., we do not require to consider all the associated events $A_{t}$'s along the stochastic path, but that we consider only the events $A_{t}$'s whose \textit{assessed} probabilities are $\alpha$. The set of those events $A_{t}%
$'s is called a \textit{test set}, because it is collected according to the selection criterion of being assessed constantly as $\alpha$. Therefore, we do not assume any specific property of the stochastic process along the path in the test set, such as stationarity or ergodicity. We do not assume any specific properties because we include only the arbitrary subsequences of the stochastic process into the test set according to the subjective assessment. 

Second, by \cref{lem:lemma2} and \cref{lem:lemma3}, we obtain that if $P(p_{k}\rightarrow$
$\alpha)=1,$ then $E$ $|\lim\limits_{k\rightarrow\infty
}\frac{1}{k}%
%TCIMACRO{\tsum \limits_{t_{j}=0}^{k-1}}%
%BeginExpansion
{\textstyle\sum\limits_{j=0}^{k-1}}
%EndExpansion
P(A_{t_{j}+1}|$\ss $_{t_{j}})-\alpha|=0$ where expectation is taken with respect to the true probability $P.$ Then, from this equation, we establish a connection between the true guarantee of well-calibration and the \textit{real} forecasting \textit{game }between a machine and Nature: $(i)$ the true guarantee of well-calibration is connected to forecasting games between a machine and Nature, for what the machine forecasts is $\alpha$ while what Nature forecasts is $P(A_{t_{j}+1}|$\ss $_{t_{j}})$ and thus whether
$| \lim\limits_{k\rightarrow\infty}\frac{1}{k}%
%TCIMACRO{\tsum \limits_{t_{j}=0}^{k-1}}%
%BeginExpansion
{\textstyle\sum\limits_{j=0}^{k-1}}
%EndExpansion
P(A_{t_{j}+1}|$\ss $_{t_{j}})-\alpha | = 0$ holds or not is tied to how Nature and the machine play in the forecasting games along the stochastic path of the test set. In this game, the machine loses at time $t$ whenever Nature succeeds in
deviating from machine forecasting at that time. There is some literature which deals with the problem of well-calibration in various forecasting game settings. (\textit{e.g.} \cite{FosterV:93}) $(ii)$ Also, note that, in the proof of \cref{lem:lemma3}, we take both the inner and outer expectations with respect to the true probability $P$ while applying the law of iterated expectations. Thus, it is a \textit{real} game, not any arbitrarily imaginary one, for $|\lim\limits_{k\rightarrow\infty}\frac{1}{k}%
%TCIMACRO{\tsum \limits_{t_{j}=0}^{k-1}}%
%BeginExpansion
{\textstyle\sum\limits_{j=0}^{k-1}}
%EndExpansion
P(A_{t_{j}+1}|$\ss $_{t_{j}})-\alpha|=0$ is expected to hold with respect to the \textit{true} probability $P$, not any other subjective probability $\Pi$.

\textbf{\cref{rem:2ndprob}} Now, let us establish a connection between the true second-order probability and the forecasting game between Nature and a machine. For simplicity, let us denote by $\Delta_{t}$ the event at time $t$ that $P(A_{t+1}|$\ss $_{t})=\alpha$ for any machine forecast $\alpha$. In other words, $\Delta_{t}$ denotes the event that the machine makes the correct forecast at time $t$, which amounts to that the machine wins the forecasting game at that time. Note here that, strictly speaking, the event $\Delta_{t}$ is a complex event which consists of two events, the event of $\{P(A_{t+1}|$\ss $_{t})=\alpha\}$ and the event of $\{\Pi(A_{t+1}|$\ss $_{t})=\alpha\}$ for the same functional
value $\alpha$ while $P(A_{t+1}|$\ss $_{t})$ and $\Pi(A_{t+1}|$\ss $_{t})$ are two probability functions about the common event $A_{t+1}$, that is $\{\Delta_{t}\}=\{P(A_{t+1}|$\ss $_{t}%
)=\alpha=\Pi(A_{t+1}|$\ss $_{t})\}$. However, since we consider only the test set along the stochastic path, here we take it that $\Pi(A_{t+1}|$\ss $_{t})$ is fixed as $\alpha$ along the path.

Then, extending some notions from \cite{Gaifman:86}, let us derive a second-order probability, i.e. the probability of probability, from the outcomes of the forecasting game between Nature and the machine as follows: for any event $A_{t+1},$ the true second-order probability $P$ is the probability of the meta-event that the first-order probability (either Nature's true forecast or the machine's subjective forecast) of $A_{t+1}$ \textit{actually} has a certain numerical value $\alpha \in \Re\lbrack0,1]$. Thus, the true second-order probability $P$ denotes $P$
$\left(\text{ }P(A_{t+1}|\text{\ss }_{t})=\alpha\text{ }\right)$.

Here, it deserves to note that although we derive the notion of higher-order probabilities by extending some notions from \cite{Gaifman:86}, our notion is different from his in the following way: we do not distinguish the first-order and the second-order probabilities while using the same notation as $\textit{P}$, although Gaifman(1986) uses $\textit{P}$ and $\textit{PR operator}$ to denote the second-order probability and the event on the first-order probability, respectively. This is because Gaifman’s notions are different from ours in that (1) $P$ in Gaifman denotes the agent $\textit{subjective}$ probability, while our second-order probability $P$ can be a true $\textit{objective}$ one just like the first-order true probability, and that (2) his $PR$ operator accepts a closed interval as one of its arguments, while our domain of the second-order probability $P$ does not contain intervals of real numbers. Note that our domain of the second-order probability is assumed to be generated by the collection of all the singletons of the computable real values of the first-order true probability function $P$, and that it is assumed to be countable. Thus, the domain does not contain intervals of real numbers. (3) In addition, our notion of the first-order probability is $\textit{not imprecise}$ but $\textit{precise}$ one, so it is not supposed to be what belongs to any interval or any set of probability measures. 

Now, the probability space of the second-order probability is defined as $(\Omega, \mathcal{G}, P)$, in which $\Omega$ is the set of all the computable functional values for any given true first-order probability function $P(A_{t+1}|\beta_{t})$, $\mathcal{G}$ is a field generated by the collection of all the singletons in $\Omega$, and $P$ is the second-order probability with $P: \mathcal{G} \rightarrow \Re\lbrack0,1]$. Note here that $\Omega$ is countable and that $\Omega$ is the set of all the possible forecasts by machines on the event $A_{t+1}$ given $\beta_t$. Now, if the domain of the second-order probability is a sigma-field $\mathcal{F}$ generated by $\Omega$, then the problem here is that the sigma-field $\mathcal{F}$ becomes uncountable given that $\Omega$ is countable. So, we should consider a field $\mathcal{G}$, not sigma-field $\mathcal{F}$ for the probability space of the second-order probability $P$.

Here are some justifications for defending the use of a field $\mathcal{G}$, not sigma-field $\mathcal{F}$, as a domain of the second-order probability $P$: we do not require the domain of the second-order probability to include all the countably infinite unions, for the number of strategies a machine can use then becomes uncountable, which is contradictory to the fact that the set of numbers a machine can compute is countable. In our forecasting game, any singleton in $\Omega$ can be thought of as a $\textit{pure}$ strategy by the machine and any union of those singletons as a $\textit{mixed}$ strategy by the machine. Again, since the set of numbers a machine can compute is countable, a machine cannot compute uncountably many mixed strategies.

\textbf{\cref{rem:tolerror}} 
Recall from the necessary condition for learning in Section 3.2 that $P(A_{t+1} | ${\ss}$_t)$ = $\Pi (A_{t+1} | ${\ss}$_t)$ = $\alpha$ if the machine learns the true probability $P(A_{t+1} | ${\ss}$_t)$ as $\alpha$. \cref{def:tolerror} then means that while the machine calculates the value of $\Pi (A_{t+1} | ${\ss}$_t)$ as $\alpha$ to learn the true probability $P(A_{t+1} | ${\ss}$_t)$ at time $t$, the machine assigns its $\Pi$- probability $>0$ to the event that $P(A_{t+1} | ${\ss}$_t) \neq \alpha$, because the machine tolerates the error that the true value of $P(A_{t+1} | ${\ss}$_t)$ may not be very $\alpha$ at that time $t$. In \cref{lem:tolerror}, we prove that a machine cannot tolerate errors infinitely often if it aims to learn the true probability.

\textbf{\cref{rem:savage}} For example, in \cite{Savage:72}, a vacuous event is null, but not every null set is necessarily vacuous. Here, an event is null to an agent when the event is \textit{believed to be} impossible to the very agent, and thus its subjective probability is zero to the agent. On the other hand, a vacuous event has absolute impossibility whose true objective probability is zero by the Kolmogorov axiom. Thus, the objective true probability of an absolutely impossible event here binds its subjective probability to zero, but not necessarily vice versa. 

We now extend this idea in \cite{Savage:72} to all virtually impossible events. Here, note that absolute impossibility is assigned to a vacuous event by the Kolmogorov axiom, while virtual impossibility is assigned to any event whose true objective probability measure is zero by Nature. Thus, in \cref{lem:tolerror}, we derive that all virtually impossible events also have a subjective probability $\Pi-$ zero infinitely often whenever the agent is self-assured that such events are truly impossible, for the subjective probability must be bound to the true objective probability $P-$ zero, if any. Otherwise, the machine comes to tolerate error infinitely often, which makes it impossible for the machine to achieve its goal of learning the true probability.

%%%%%%%%%%%%%%%%%%%%%%%%%%%%%%%%%%%%%%%%%%%%%%%%%%%%%%%%%%%%%%%%%%%%%%%%%%%%%%%
%%%%%%%%%%%%%%%%%%%%%%%%%%%%%%%%%%%%%%%%%%%%%%%%%%%%%%%%%%%%%%%%%%%%%%%%%%%%%%%

\end{document}